\documentclass{article} % For LaTeX2e
\usepackage[OT1]{fontenc}
\usepackage{arxiv}
\usepackage{natbib}
\setcitestyle{authoryear,round,citesep={;},aysep={,},yysep={;}}
\usepackage{xcolor}

\usepackage{amsmath,amsfonts,bm}

\def\eqref#1{equation~\ref{#1}}
\def\1{\bm{1}}

\DeclareMathAlphabet{\mathsfit}{\encodingdefault}{\sfdefault}{m}{sl}
\SetMathAlphabet{\mathsfit}{bold}{\encodingdefault}{\sfdefault}{bx}{n}

\usepackage{hyperref}
\usepackage{url}
\hypersetup{
    hidelinks,
    pdftitle={Visual Autoregressive Generation with Next Token Imagination},
    pdfauthor={Ruihang Jiang, Zongyao Zhao, Zhaolin Wang, Yuanwei Liu, Xiatian Zhu}
}

\usepackage{graphicx}
\usepackage{multirow}
\usepackage{float}
\usepackage{booktabs}
\usepackage{caption}
\usepackage{tabularx}
\usepackage{wrapfig}

\usepackage{array}
\newcolumntype{Y}{>{\raggedright\arraybackslash}X}

\definecolor{NTILink}{RGB}{36,95,140}

\title{Visual Autoregressive Generation \\ with Next Token Imagination}

\author{\bfseries Ruihang Jiang$^{1}$ \quad Zongyao Zhao$^{1}$ \quad
Zhaolin Wang$^{1}$ \quad Yuanwei Liu$^{1}$ \quad Xiatian Zhu$^{2}$ \\[5pt]
\normalfont $^{1}$University of Hong Kong \qquad $^{2}$University of Surrey}

\date{}
\renewcommand{\shorttitle}{Next Token Imagination}
\begin{document}

\maketitle

\begin{abstract}

The space of possible visual content is enormous, and arguably unbounded. Any training set, however large, therefore covers only a tiny fraction of it. Yet generative models are typically trained to fit only the images they are given. In visual autoregressive (AR) generation, for example, each token is predicted from the preceding ground-truth (GT) tokens and supervised solely by the GT next token. Confining learning to observed data in this way limits how much of the visual distribution a model can capture. In this work, we hypothesize that letting a model \textit{imagine} beyond its training data, by exploring plausible visual realizations it has never observed, can ease this limitation. We use visual AR generation as a proof of concept, since its fixed, discrete token space makes imagination easier to steer and constrain. We propose \textbf{Next Token Imagination (NTI)}, a method that allows the model to explore beyond the GT next token, subject to a carefully designed criterion that keeps this imagination bounded by the underlying visual distribution. Extensive experiments on ImageNet show that NTI substantially and consistently improves the fidelity of generated images, approaching state-of-the-art diffusion and AR models while using a smaller model and lower training cost.
Code and checkpoints are available at \href{https://github.com/JackRuihang/NTI}{\textcolor{NTILink}{https://github.com/JackRuihang/NTI}}.
\end{abstract}

\section{Introduction}

Autoregressive (AR) models, trained to predict each token from its preceding context, form the basis of modern language generation \citep{brown2020language}. Inspired by their success in text, AR modeling has been extended to image generation through discrete visual tokenization \citep{esser2021taming,sun2024autoregressive}. However, standard visual AR models still lag behind leading diffusion and flow-matching models in generation quality \citep{peebles2023scalable,ma2024sit,yu2025repa}. We therefore revisit visual AR training to identify what constrains its generative capability.

The standard visual AR pipeline consists of a visual tokenizer and a causal Transformer. The tokenizer encodes each image into a sequence of discrete codebook indices and decodes the corresponding codebook embeddings to reconstruct the image \citep{oord2017neural,esser2021taming,lee2022autoregressive}. The Transformer models this sequence, typically in raster order, through next token prediction \citep{parmar2018image,esser2021taming,sun2024autoregressive}. Advances in visual AR have largely focused on tokenizer design \citep{yu2024language,xiong2025gigatok}, token representation \citep{yu2024image,bachmann2025flextok}, generation order \citep{yu2025randomized,pang2025randar}, and model scaling \citep{sun2024autoregressive,han2025infinity}. In contrast, generator training still follows the standard paradigm. During training, each position is supervised by the corresponding ground-truth (GT) next token, with the prediction conditioned on the GT prefix \citep{bengio2015scheduled,sun2024autoregressive}. At inference, however, the model samples a token from its predictive distribution at each position, so non-GT tokens can accumulate through AR generation, causing the generated sequence to depart from the GT sequence.

In this work, we revisit standard visual AR training from a distributional perspective. A non-GT next token does not necessarily move the generation outside the space of plausible visual realizations. Instead, it can lead subsequent generation toward another valid visual realization beyond the GT sequence. Through controlled experiments, we examine this hypothesis, motivating the model to explore plausible non-GT next tokens during training.

\begin{figure}[t]
    \centering
    \setlength{\parskip}{0pt}

    \includegraphics[width=\linewidth]{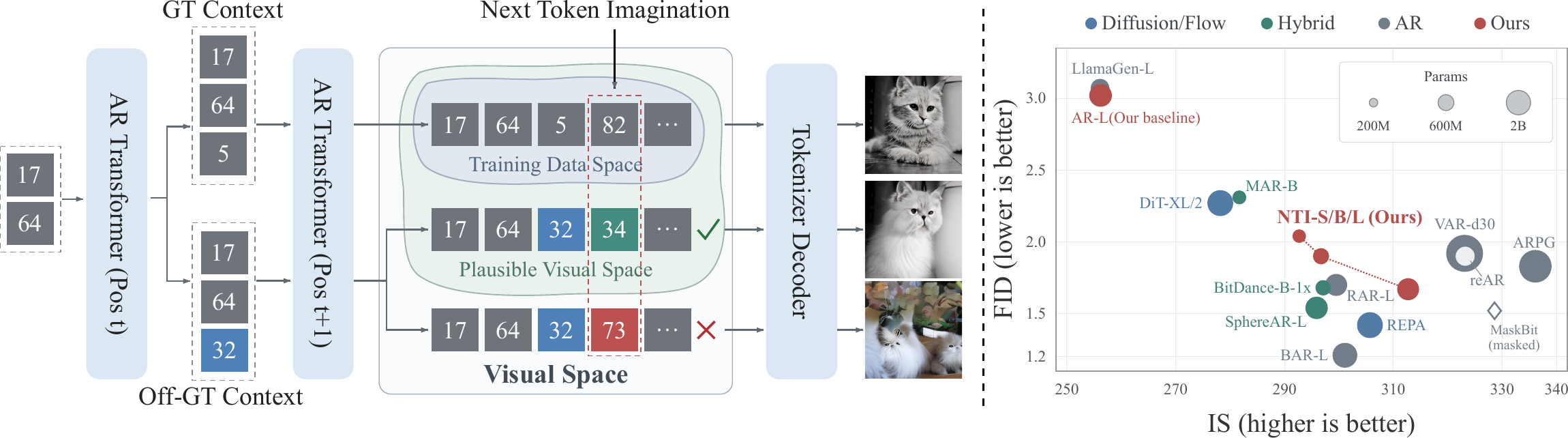}\par
    \vspace{4pt}
    \nointerlineskip
    \noindent
    \begin{minipage}[t]{0.65\linewidth}
        \small
        (a) NTI breaks the GT-only constraint in visual AR training by expanding next token prediction to plausible non-GT choices. These choices naturally induce off-GT contexts, where NTI regulates subsequent imagination to favor plausible continuations and explore visual space beyond observed GT sequences.
    \end{minipage}\hfill
    \begin{minipage}[t]{0.32\linewidth}
        \small
        (b) NTI substantially advances standard raster-order AR generation on ImageNet, reducing FID from 3.02 to 1.67 and increasing IS from 256.2 to 312.72.
    \end{minipage}\par
    \caption{\textbf{NTI unlocks broader visual space for stronger visual AR generation.}}
    \label{fig:intro_merge}
    \vspace{-8pt}
\end{figure}

First, standard visual AR training imposes a \textbf{restricted next token space}. Given a context, the GT next token represents only one observed choice within a much larger space of plausible continuations. As illustrated in Figure~\ref{fig:intro_merge}(a), the generator predicts token 32 instead of the GT token 5, while the resulting sequence still produces a visually plausible cat. This indicates that plausible non-GT next tokens exist beyond the GT target, motivating the model to explore a broader next token space during training.

Second, a \textbf{restricted next token space} further leads to a \textbf{restricted context space} during training. Since each sampled token becomes part of the prefix for later positions, sampling a non-GT token naturally creates an off-GT context. Related work on exposure bias \citep{bengio2015scheduled,wang2020exposure} typically treats such contexts as consequences of prediction errors and focuses on improving model robustness. We instead view off-GT contexts as natural states induced by the model's next token choices. The same example in Figure~\ref{fig:intro_merge}(a) further illustrates this process. Once the model samples token 32 instead of the GT token 5, the next position is conditioned on an off-GT context. Given this prefix, token 34 still produces a visually plausible cat, whereas token 73 leads to structural distortion. This suggests that broadening the next token space alone is insufficient; in other words, training must distinguish plausible non-GT predictions from implausible ones.

Motivated by these observations, we propose \textbf{Next Token Imagination (NTI)}, a plug-and-play training paradigm that breaks the constraint of GT-only next token supervision. Specifically, NTI uses \textbf{Context-Regulated Imagination (CRI)} to assign probability to plausible non-GT tokens according to the current context. Ideally, such imagination should be incorporated directly into the AR process, where each imagined next token becomes part of the context for subsequent positions. However, performing this sequential rollout during training is extremely expensive. We therefore introduce \textbf{Snapshot-based Context Approximation (SCA)} to efficiently construct model-generated contexts without explicit sequential rollout. Together, these two mechanisms move visual AR training beyond the observed GT sequence while keeping imagination regulated.

Experiments on ImageNet establish NTI as a strong training paradigm for visual AR generation. With a standard raster-order AR backbone, NTI reduces FID from $3.02$ to $1.67$ and improves IS from $256.20$ to $312.72$, with consistent gains across model scales. NTI further outperforms representative AR methods including reAR~\citep{he2026rear}, RAR~\citep{yu2025randomized}, LlamaGen~\citep{sun2024autoregressive}, and BitDance~\citep{ai2026bitdance}, while remaining competitive with substantially larger or more specialized generative models. NTI also transfers effectively to non-standard tokenizers such as GigaTok~\citep{xiong2025gigatok} and AliTok~\citep{wu2025alitok}, demonstrating that its gains are not tied to a particular tokenizer design. Notably, these improvements are achieved without changing the inference pipeline. Collectively, these results establish NTI as a general and effective training paradigm for substantially advancing visual AR generation.

Our contributions are threefold. \textbf{(1)} We expose a fundamental limitation of visual AR training: GT-only next token supervision unnecessarily restricts valid continuations beyond the observed GT sequence. \textbf{(2)} We introduce \textbf{NTI}, a plug-and-play training paradigm that enables context-regulated exploration beyond the GT sequence. \textbf{(3)} We demonstrate that NTI consistently improves visual AR generation across model scales and tokenizer designs, without modifying the tokenizer, generation order, or inference pipeline.

\begin{figure}[t]
    \centering
    \setlength{\parskip}{0pt}

    \includegraphics[width=\linewidth]{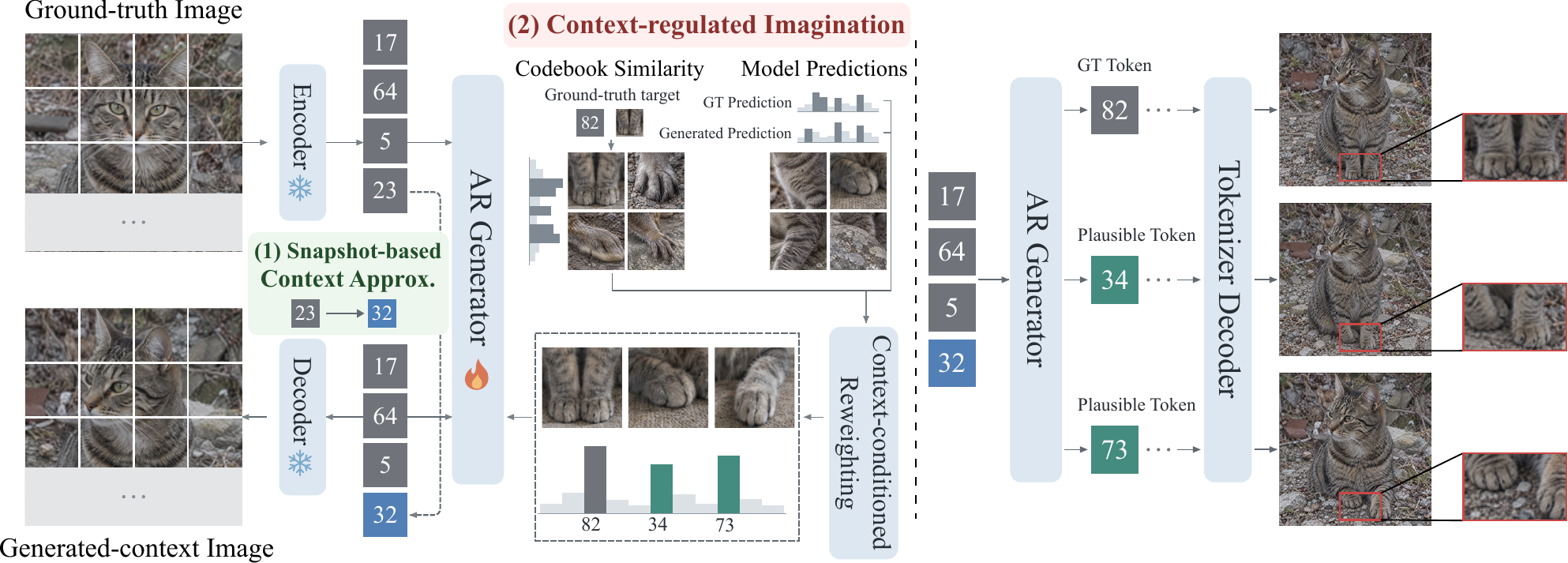}\par
    \vspace{4pt}
    \nointerlineskip
    \noindent
    \begin{minipage}[t]{0.65\linewidth}
        \centering
        \small
        (a) \textbf{NTI Training: SCA + CRI}
    \end{minipage}\hfill
    \begin{minipage}[t]{0.32\linewidth}
        \centering
        \small
        (b) \textbf{NTI Inference}
    \end{minipage}\par
    \caption{\textbf{Overview of NTI, a plug-and-play training paradigm with regulated imagination.}}
    \label{fig:riar_overview}
\end{figure}

\section{Preliminaries}
\label{sec:preliminaries}

Visual autoregressive generation follows a three-stage pipeline. (1) A visual tokenizer encodes images into latent sequences. (2) An AR generator models the distribution over these sequences and generates new sequences at inference. (3) The generated sequences are decoded back into images by the tokenizer decoder.

\paragraph{Visual tokenizer.}
Visual AR models commonly use discrete visual tokenizers to represent images as codebook indices \citep{oord2017neural,esser2021taming,lee2022autoregressive}. The tokenizer consists of an encoder $\mathcal{E}$, a vector quantizer $\mathcal{Q}$ with a codebook $\mathcal{C}=\{\mathbf{c}_k\}_{k=1}^{K}\subset\mathbb{R}^{d}$, and a decoder $\mathcal{D}$, where $K$ is the codebook size and $d$ is the latent embedding dimension. Given an image $\mathbf{I}\in\mathbb{R}^{3\times H\times W}$, where $H$ and $W$ denote the image height and width, the encoder maps it to a latent feature grid $\mathbf{u}=\mathcal{E}(\mathbf{I})\in\mathbb{R}^{h\times w\times d}$, where $h$ and $w$ denote the height and width of the latent grid. The quantizer $\mathcal{Q}$ assigns each latent feature to its nearest codebook entry, forming a quantized latent grid that is decoded by $\mathcal{D}$ to reconstruct the image:
\begin{equation}
z_{ij}=\arg\min_{k\in\{1,\ldots,K\}}
\left\lVert \mathbf{u}_{ij}-\mathbf{c}_k \right\rVert_2^2,
\qquad
\widehat{\mathbf{u}}_{ij}=\mathbf{c}_{z_{ij}},
\qquad
\widehat{\mathbf{I}}=\mathcal{D}(\widehat{\mathbf{u}}).
\label{eq:visual_tokenization}
\end{equation}

\paragraph{Visual autoregressive generation.}

In standard visual AR models, the $h\times w$ grid of codebook indices is typically flattened in raster order into a token sequence $\mathbf{z}=(z_1,\ldots,z_N)$ of length $N=hw$ \citep{parmar2018image,esser2021taming,sun2024autoregressive}. We refer to $\mathbf{z}$ as the \textbf{GT sequence}. Given conditioning information $y$, a causal generator $p_{\theta}$ factorizes its distribution as
\begin{equation}
p_{\theta}(\mathbf{z}\mid y)
=
\prod_{t=1}^{N}
p_{\theta}(z_t\mid\mathbf{z}_{<t},y),
\qquad
\mathbf{z}_{<t}=(z_1,\ldots,z_{t-1}).
\label{eq:ar_factorization}
\end{equation}

We refer to $\mathbf{z}_{<t}$ as the \textbf{GT context} at position $t$. During standard training, the generator predicts the \textbf{GT next token} $z_t$ from $\mathbf{z}_{<t}$ by minimizing the negative log-likelihood:
\begin{equation}
\mathcal{L}_{\mathrm{AR}}(\theta)
=
-\frac{1}{N}\sum_{t=1}^{N}
\log p_{\theta}(z_t\mid\mathbf{z}_{<t},y).
\label{eq:ar_training}
\end{equation}

At inference, the generator instead conditions on its previously generated tokens. We denote the generated token sequence as $\widetilde{\mathbf{z}}=(\widetilde{z}_1,\ldots,\widetilde{z}_N)$ and its prefix $\widetilde{\mathbf{z}}_{<t}=(\widetilde{z}_1,\ldots,\widetilde{z}_{t-1})$ as the \textbf{generated context} at position $t$:
\begin{equation}
\widetilde{z}_t
\sim
p_{\theta}\bigl(\cdot\mid\widetilde{\mathbf{z}}_{<t},y\bigr).
\label{eq:ar_inference}
\end{equation}

\section{Next Token Imagination}
\label{sec:method}

We formulate visual AR training from the perspective of next token prediction. At position $t$, standard training predicts the GT next token $z_t$ conditioned on the GT context $\mathbf{z}_{<t}$. At inference, the sampled token $\widetilde{z}_t$ becomes part of the context for the next position, with $\widetilde{\mathbf{z}}_{<t+1}=(\widetilde{\mathbf{z}}_{<t},\widetilde{z}_t)$. Existing methods on exposure bias introduce such off-GT contexts during training but continue to supervise each position with the GT token $z_t$. Such supervision implicitly regards off-GT contexts as states arising from accumulated prediction errors and trains the model to return toward the GT sequence. This view is natural for conventional compression, where the objective is to recover a given source. Generation is different: the plausible visual space is enormous, while the GT sequences in the training data cover only a small fraction of it. Forcing every off-GT context back toward the GT sequence therefore suppresses the diversity that a generative model should capture. Valid non-GT next token predictions should instead be encouraged, allowing the model to recursively explore valid continuations beyond the observed GT sequences and thereby cover a broader plausible visual space.

We quantitatively analyze the effects of deviations in next token predictions and their induced contexts on visual generation in Section~\ref{sec:understanding}. Based on these results, we propose \textbf{NTI}, a plug-and-play training paradigm for visual AR models. Specifically, \textbf{CRI} extends the supervision from the GT token to plausible non-GT ones. To improve training efficiency, instead of autoregressively rolling out contexts, we introduce \textbf{SCA} to construct model-generated contexts in a single forward pass. Together, they allow the model to explore beyond the visual space covered by the training data.

\subsection{Understanding Restricted Next Token Space}
\label{sec:understanding}

Given a GT context $\mathbf{z}_{<t}$ and a generated context $\widetilde{\mathbf{z}}_{<t}$, we consider a context window $\mathcal{W}_t\subseteq\{1,\ldots,t-1\}$ of length $L=|\mathcal{W}_t|$. Let $\mathcal{A}=\{i\in\mathcal{W}_t:\widetilde{z}_i\neq z_i\}$ denote the positions where the two contexts differ. We measure their deviation using \textbf{Ground Truth Token Ratio (GTR)} and \textbf{Ground Truth Token Similarity (GTS)}; specifically, $\operatorname{GTR}=1-|\mathcal{A}|/L$ and $\operatorname{GTS}=|\mathcal{A}|^{-1}\sum_{i\in\mathcal{A}}\operatorname{sim}(\mathbf{c}_{\widetilde{z}_i},\mathbf{c}_{z_i})$, with $\operatorname{GTS}=1$ when $\mathcal{A}=\emptyset$, where $\operatorname{sim}(\cdot,\cdot)$ denotes cosine similarity. We further use LPIPS~\citep{zhang2018unreasonable} and a class-conditional DINOv2 similarity score~\citep{oquab2024dinov2} to evaluate perceptual deviation and class-conditional semantic similarity in the decoded images. The following two controlled experiments show that NTI is feasible but requires regulation.

\begin{figure}[t]
    \centering
    \setlength{\parskip}{0pt}

    \includegraphics[width=\linewidth]{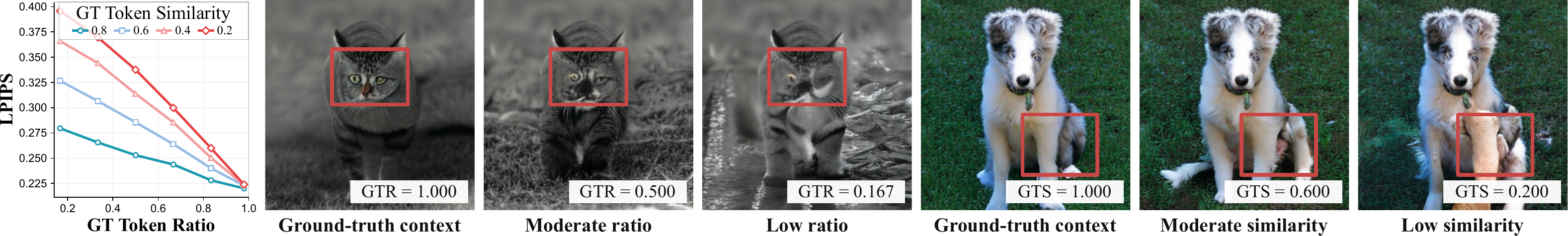}\par
    \vspace{4pt}
    \nointerlineskip

    \caption{\textbf{NTI is feasible beyond the GT sequence: off-GT contexts can remain within a plausible visual space.}}
    \label{fig:experiment1}
\end{figure}

\begin{figure}[t]
    \centering
    \setlength{\parskip}{0pt}

    \includegraphics[width=\linewidth]{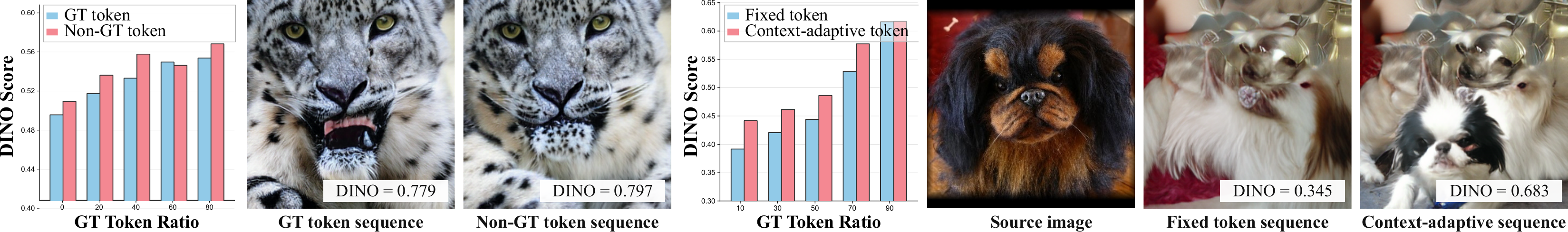}\par
    \vspace{4pt}
    \nointerlineskip
    \noindent
    \begin{minipage}[t]{0.49\linewidth}
        \centering
        \small
        (a) \textbf{GT vs.\ non-GT tokens}
    \end{minipage}\hfill
    \begin{minipage}[t]{0.49\linewidth}
        \centering
        \small
        (b) \textbf{Fixed vs.\ context-adaptive targets}
    \end{minipage}\par
    \vspace{0pt}

    \caption{\textbf{NTI requires context regulation: non-GT tokens can be plausible supervision targets, but their validity depends on the current context.}}
    \label{fig:experiment2}
\end{figure}

\noindent\textbf{Feasibility of NTI.}
To examine whether NTI can move beyond the GT sequence while remaining within a plausible visual space, we study how contexts induced by non-GT next token predictions affect subsequent generation. We construct perturbed contexts from GT contexts under two settings: varying GTR at fixed GTS, and varying GTS while fixing the replacement positions and GTR. Starting from each perturbed context, we autoregressively generate the remaining tokens and measure the perceptual deviation of the decoded images using LPIPS.

Figure~\ref{fig:experiment1} shows that context deviations influence subsequent generation: lower GTR and lower GTS generally lead to higher LPIPS, indicating larger perceptual deviations in the decoded images. However, off-GT contexts do not necessarily result in failed generations. As shown in the cat examples, the facial features become progressively blurred as the context deviates further from the GT context, while in the dog examples, moderate deviations still preserve coherent leg structures before severe distortions emerge. These results demonstrate that contexts induced by non-GT next token predictions can remain within a plausible visual space, supporting NTI beyond the GT sequence.

\noindent\textbf{Regulation of NTI.}
Having established that NTI can move beyond the GT sequence, we further investigate how NTI should be regulated. Specifically, we examine \textbf{(1)} whether non-GT tokens can serve as plausible supervision targets under off-GT contexts, and \textbf{(2)} whether suitable targets vary with the current context. We conduct two comparisons under generated contexts and evaluate the resulting images using DINO score.

For the first aspect, we compare GT and non-GT tokens under the same generated context. Across different GTR values, non-GT tokens achieve comparable or even higher mean DINO scores than GT tokens. Figure~\ref{fig:experiment2}(a) shows this trend, while the snow leopard examples demonstrate that both choices can produce visually plausible images. These results demonstrate that the GT token is not the only plausible supervision target under an off-GT context, supporting NTI beyond the GT token.

Based on this observation, a straightforward approach is to include non-GT tokens according to codebook similarity~\citep{hu2025improving,tang2025exploiting}. However, such targets remain fixed across different contexts. We therefore compare fixed selection, which uses the same non-GT token across contexts, with context-adaptive selection. As shown in Figure~\ref{fig:experiment2}(b), the two perform similarly at high GTR, while context-adaptive selection performs better as GTR decreases. When GTR decreases further, both non-GT choices degrade, indicating that the imagination range should contract toward the GT token. These results demonstrate that NTI must be regulated by the current context.

\subsection{Training with Next Token Imagination}
\label{sec:nti_training}

We realize NTI with a simple and efficient training procedure. \textbf{SCA} constructs model-generated contexts without AR rollout. Using predictions under the GT and model-generated contexts, \textbf{CRI} directly determines which non-GT tokens should be included in supervision and how much probability they should receive.

\noindent\textbf{Snapshot-based Context Approximation.}
Based on the above findings, the ideal NTI training procedure would construct each context and its corresponding supervision through step-by-step AR rollout. However, performing AR rollout for every sequence is computationally expensive. Previous work uses scheduled sampling~\citep{bengio2015scheduled,zhan2022autoregressive} to construct model-generated contexts. Such online construction, however, requires an additional prediction pass during training. We therefore introduce \textbf{SCA}, which uses a periodically refreshed frozen snapshot $p_{\bar{\theta}}$ of the current generator $p_\theta$ to approximate model-generated contexts for NTI while preserving a single forward pass during training.

Concretely, given a GT sequence $\mathbf{z}$, the snapshot computes the prediction distribution under the GT context as $\mathbf{p}_t^{\mathrm{GT}}=p_{\bar{\theta}}(\cdot\mid\mathbf{z}_{<t},y)$ for all positions in parallel. We sample $\widehat{z}_i\sim\mathbf{p}_i^{\mathrm{GT}}$ and $b_i\sim\operatorname{Bernoulli}(\rho)$, where $\rho$ controls the probability of replacing a GT token with a model prediction. We then construct $\widetilde{\mathbf{z}}$ as
\begin{equation}
\widetilde{z}_i=
\begin{cases}
z_i, & b_i=0,\\
\widehat{z}_i, & b_i=1,
\end{cases}
\qquad i=1,\ldots,N.
\label{eq:context_construction}
\end{equation}

For each GT sequence, we sample $\rho\sim\mathcal{U}(0,\rho_{\max}(s))$, where $s$ denotes training progress and $\rho_{\max}(s)$ controls the maximum replacement ratio. The prefix $\widetilde{\mathbf{z}}_{<t}$ approximates the model-generated context at position $t$, and the snapshot computes $\mathbf{p}_t^{\mathrm{gen}}=p_{\bar{\theta}}(\cdot\mid\widetilde{\mathbf{z}}_{<t},y)$ from this context. The predictions under the GT and model-generated contexts are then used to regulate NTI.

\noindent\textbf{Context-Regulated Imagination.}
To construct the NTI supervision target, we address three questions: \textbf{(1)} which non-GT tokens can be included in imagination; \textbf{(2)} how their relative weights should adapt to the current context; and \textbf{(3)} how much supervision should be assigned to imagination overall. We answer these questions using codebook similarity, the predictions under the GT and model-generated contexts, and model confidence, respectively.

\noindent\textit{(1) Imagination Candidates.}
To determine which non-GT tokens can be included in imagination, we first use codebook similarity as a context-independent prior. Our analysis in Section~\ref{sec:understanding} shows that replacing GT tokens with more similar codebook entries generally induces smaller perceptual deviations, suggesting that nearby tokens are more likely to remain within the plausible visual space. We therefore define the affinity between each token $k$ and the GT token $z_t$ as
\begin{equation}
A_{z_t,k}
=
\exp\left(
\frac{\operatorname{sim}(\mathbf{c}_{z_t},\mathbf{c}_k)-1}{\tau}
\right),
\label{eq:token_affinity}
\end{equation}
where $\operatorname{sim}(\cdot,\cdot)$ denotes cosine similarity and $\tau>0$ controls the concentration around the GT token. This affinity identifies plausible non-GT candidates for imagination, but remains fixed across different contexts.

\noindent\textit{(2) Context-Adaptive Weighting.}
Our analysis in Figure~\ref{fig:experiment2}(b) shows that suitable non-GT targets depend on the current context. Codebook affinity alone cannot capture this dependency. We therefore compare the prediction distributions under the GT and model-generated contexts, $\mathbf{p}_t^{\mathrm{GT}}$ and $\mathbf{p}_t^{\mathrm{gen}}$, and measure their consistency using total variation (TV) distance:
\begin{equation}
g_t
=
1-\operatorname{TV}
\left(
\mathbf{p}_t^{\mathrm{GT}},
\mathbf{p}_t^{\mathrm{gen}}
\right)
=
1-\frac{1}{2}
\sum_{k=1}^{K}
\left|
\mathbf{p}_t^{\mathrm{GT}}(k)
-
\mathbf{p}_t^{\mathrm{gen}}(k)
\right|,
\label{eq:context_overlap}
\end{equation}
where $K$ is the codebook size. A larger $g_t$ indicates more consistent predictions under the two contexts. We combine the two distributions according to $g_t$, relying more on the model-generated context when they remain consistent and shifting toward the GT context as they diverge. We then combine this context-dependent prediction with the codebook affinity and normalize over non-GT tokens to obtain $r_t$, a distribution over non-GT tokens:
\begin{equation}
\begin{aligned}
u_t(k)
&=
A_{z_t,k}
\left[
(1-g_t)\mathbf{p}_t^{\mathrm{GT}}(k)
+
g_t\mathbf{p}_t^{\mathrm{gen}}(k)
\right],\\
r_t(k)
&=
\begin{cases}
\displaystyle
\frac{u_t(k)}
{\sum_{j\neq z_t}u_t(j)},
&
k\neq z_t,\\
0,
&
k=z_t.
\end{cases}
\end{aligned}
\label{eq:alternative_distribution}
\end{equation}

\noindent\textit{(3) Imagination Strength.}
Finally, we determine how much supervision should be assigned to imagination. While $r_t$ specifies the distribution over non-GT tokens, we introduce $d_t$ to control their total probability. We compute $d_t$ from three factors. First, $c_t=1-\mathcal{H}(\mathbf{p}_t^{\mathrm{GT}})/\log K$ measures the prediction confidence under the GT context, where $\mathcal{H}$ denotes Shannon entropy. Second, $1-\mathbf{p}_t^{\mathrm{GT}}(z_t)$ measures how much probability the model assigns beyond the GT token. Third, $g_t$ measures the consistency between predictions under the GT and model-generated contexts. Combining these factors, we define
\begin{equation}
d_t
=
c_t
\bigl(1-\mathbf{p}_t^{\mathrm{GT}}(z_t)\bigr)
g_t,
\end{equation}
so that imagination is reduced when the model is uncertain, strongly favors the GT token, or becomes inconsistent under the model-generated context. The final NTI target distribution is
\begin{equation}
q_t(k)
=
(1-d_t)\mathbf{1}\{k=z_t\}
+
d_t r_t(k).
\label{eq:adaptive_supervision}
\end{equation}

\noindent\textbf{Training Objective.}
NTI augments the standard AR objective with context-regulated supervision under model-generated contexts. We keep $\mathcal{L}_{\mathrm{AR}}$ in Equation~\ref{eq:ar_training} for direct learning from the GT sequence. For an SCA sequence $\widetilde{\mathbf{z}}$, let $m_t=\mathbf{1}[\widetilde{\mathbf{z}}_{<t}\neq\mathbf{z}_{<t}]$ indicate whether position $t$ has a model-generated context. We supervise positions with $m_t=1$ using the NTI target distribution $q_t$:
\begin{equation}
\mathcal{L}_{\mathrm{NTI}}(\theta)
=
-
\frac{1}{
\mathbb{E}_{\mathrm{SCA}}
\left[
\sum_{t=1}^{N} m_t
\right]
}
\,
\mathbb{E}_{\mathrm{SCA}}
\left[
\sum_{t=1}^{N}
m_t
\sum_{k=1}^{K}
q_t(k)
\log p_\theta
\left(
k\mid\widetilde{\mathbf{z}}_{<t},y
\right)
\right].
\label{eq:nti_loss}
\end{equation}

GT and model-generated sequences are mixed within each training batch and processed in a single forward pass. The AR and NTI objectives are normalized over their respective supervised tokens, with $\lambda(s)=\lambda\rho_{\max}(s)/\rho_{\max}$ controlling the contribution of NTI supervision, and combined as
\begin{equation}
\mathcal{L}(\theta)
=
\mathcal{L}_{\mathrm{AR}}(\theta)
+
\lambda(s)\mathcal{L}_{\mathrm{NTI}}(\theta).
\label{eq:nti_objective}
\end{equation}

\begin{table}[t]
\centering
\caption{\textbf{Results on $256\times256$ class-conditional generation on ImageNet-1K.}
N/A indicates that a discrete codebook is not applicable; N/R denotes values not reported in the original paper; -- indicates that the corresponding checkpoint is not publicly available.
Throughput is measured in images per second.
$\uparrow$ and $\downarrow$ indicate that higher and lower values are better, respectively.
$^{\dagger}$ indicates that the model is not provided and it\textquoteright{}s trained with our implementation.}
\label{tab:imagenet_comparison}

\vspace{4pt}

\begingroup
\normalfont
\mdseries
\fontsize{7.8}{9.6}\selectfont
\setlength{\tabcolsep}{2.2pt}
\setlength{\arrayrulewidth}{0.25pt}
\renewcommand{\arraystretch}{1.20}

\begin{tabular}{@{}llccrrrr@{}}
\hline
Type
& Model
& Codebook size
& Epochs
& \#Params
& FID$\downarrow$
& IS$\uparrow$
& Throughput (img/s)$\uparrow$ \\
\hline

% ---------------- Diffusion / Flow ----------------
\multirow{3}{*}{Diff./Flow}
& DiT-XL/2 \citep{peebles2023scalable}
& N/A & 1,400 & 675M & 2.27 & 278.2 & 0.592 \\

& REPA \citep{yu2025repa}
& N/A & 800 & 675M & 1.42 & 305.7 & 0.671 \\

& FreqFlow-H \citep{ren2026frequency}
& N/A & N/R & 1.08B & 1.38 & N/R & -- \\
\hline

% ---------------- Masked Generation ----------------
\multirow{2}{*}{Masked Gen.}
& MAR-B \citep{li2024autoregressive}
& N/A & 800 & 208M & 2.31 & 281.7 & 0.541 \\

& MaskBit \citep{weber2024maskbit}
& $16{,}384$ & 1,080 & 305M & 1.52 & 328.6 & 0.949 \\
\hline

% ---------------- Alternative-order AR ----------------
\multirow{3}{*}{Alt.-order AR}
& VAR-d30 \citep{tian2024visual}
& $4{,}096$ & 350 & 2.0B & 1.92 & 323.1 & 26.759 \\

& RAR-L \citep{yu2025randomized}
& $1{,}024$ & 400 & 461M & 1.70 & 299.5 & 12.864 \\

& ARPG-XXL (32 steps) \citep{li2026autoregressive}
& $16{,}384$ & 400 & 1.3B & 1.83 & 336.1 & 43.861 \\
\hline

% ---------------- Continuous AR ----------------
\multirow{2}{*}{Continuous AR}
& SphereAR-L \citep{ke2026hyperspherical}
& N/A & 400 & 479M & 1.54 & 295.9 & 2.375 \\

& BAR-L \citep{tang2026bar}
& N/A & 800 & 608M & 1.21 & 301.1 & -- \\
\hline

% ---------------- Sequential AR ----------------
\multirow{3}{*}{Sequential AR}
& LlamaGen-L \citep{sun2024autoregressive}
& $16{,}384$ & 300 & 343M & 3.07 & 256.1 & 2.194 \\

& BitDance-B-1x \citep{ai2026bitdance}
& $2^{32}$ & 800 & 242M & 1.68 & 297.1 & 1.701 \\

& reAR-L (CFG=$11$) \citep{he2026rear}
& $1{,}024$ & 400 & 461M
& 1.90
& 323.2
& 9.522 \\
\hline

% ---------------- Ours ----------------
\multirow{4}{*}{Ours}
& AR-L$^{\dagger}$ \citep{yu2025randomized}
& $1{,}024$ & 400 & 461M
& 3.02
& 256.20
& -- \\

& \hspace{4pt}\textbf{NTI-S}
& $1{,}024$ & 400 & 218M
& ${}^{\pm 0.03}2.04$
& ${}^{\pm 1.41}292.70$
& 20.709 \\

& \hspace{4pt}\textbf{NTI-B}
& $1{,}024$ & 400 & 261M
& ${}^{\pm 0.02}1.90$
& ${}^{\pm 0.76}296.71$
& 17.519 \\

& \hspace{4pt}\textbf{NTI-L} (CFG=14)
& $1{,}024$ & 400 & 461M
& ${}^{\pm 0.01}1.67$
& ${}^{\pm 1.02}312.72$
& 13.957 \\
\hline

\end{tabular}

\endgroup
\end{table}

\section{Experiments}
\label{sec:experiments}

\subsection{Experimental Setup}
\label{sec:experimental_setup}

\noindent\textbf{Dataset and evaluation.}
We conduct class-conditional generation experiments on ImageNet-1K~\citep{deng2009imagenet} at a resolution of $256\times256$. Following the ADM evaluation protocol~\citep{dhariwal2021diffusion}, we generate $50{,}000$ images with classifier-free guidance~\citep{ho2022classifierfree} and report Fr\'echet Inception Distance (FID)~\citep{heusel2017gans} and Inception Score (IS)~\citep{salimans2016improved}.

\noindent\textbf{Model configuration.}
We adopt a pretrained MaskGIT-VQGAN tokenizer~\citep{chang2022maskgit} with $1{,}024$ codebook entries and a spatial downsampling factor of $16$, encoding each image into $256$ discrete tokens. NTI-S/B/L consist of $20/24/24$ causal Transformer layers with hidden dimensions of $768/768/1024$, respectively, and all variants use $16$ attention heads.

\noindent\textbf{Training.}
We train all models for $400$ epochs with a global batch size of $2{,}048$, using AdamW~\citep{loshchilov2019decoupled}. The learning rate is linearly warmed up to $4\times10^{-4}$ during the first $100$ epochs and then decayed to $10^{-5}$ with a cosine schedule over the remaining $300$ epochs. We apply gradient clipping with a maximum norm of $1.0$ and drop class labels with probability $0.1$ for classifier-free guidance~\citep{ho2022classifierfree}.

\noindent\textbf{NTI implementation.}
We set the affinity temperature to $\tau=0.05$. The maximum replacement probability first increases to $0.3$ and then decreases to zero by epoch $320$, after which the remaining $80$ epochs use only GT sequences. SCA contexts are regenerated every $16$ epochs. Further experimental details are provided in Appendix~\ref{app:experimental_details}.

% ---------------------------------------------------------
\begin{figure*}[t]
    \centering

    % ==================== Left: Scaling ====================
    \begin{minipage}[t]{0.45\textwidth}
        \vspace{0pt}
        \centering
        \includegraphics[width=0.89\linewidth]{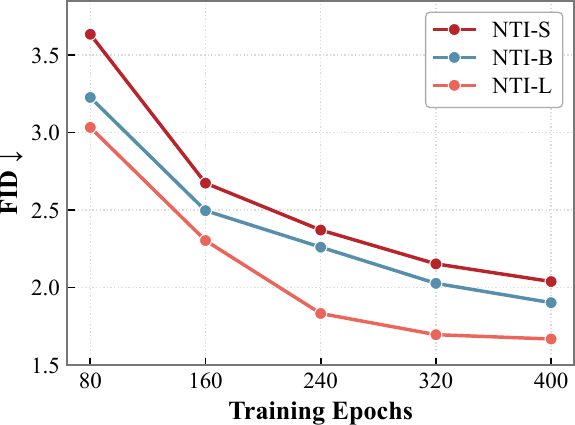}

        \captionof{figure}{
        \textbf{Scaling effect of NTI.}
        }
        \label{fig:scaling_effect}
    \end{minipage}
    \hfill
    % ================= Right: Generalization ===============
    \begin{minipage}[t]{0.54\textwidth}
        \vspace{0pt}
        \centering

        \captionof{table}{
        \textbf{Generalization of NTI.}
        NTI consistently improves non-standard visual AR models without changing their tokenizer or generator.
        }
        \label{tab:generalization}

        \vspace{0.4em}

        \small
        \setlength{\tabcolsep}{5pt}
        \begin{tabular}{lccc}
            \toprule
            Model & Epochs & Params. & FID $\downarrow$ \\
            \midrule

            \multicolumn{4}{l}{\textit{GigaTok}} \\
            GigaTok baseline & 400 & 111M & 3.94 \\
            GigaTok + NTI & 400 & 111M & \textbf{3.66} \\

            \midrule

            \multicolumn{4}{l}{\textit{AliTok}} \\
            AliTok baseline & 400 & 177M & 1.61 \\
            AliTok + NTI & 400 & 177M & \textbf{1.51} \\

            \bottomrule
        \end{tabular}
    \end{minipage}

    \vspace{-6pt}
\end{figure*}
% ---------------------------------------------------------

\subsection{Main Results}
\label{sec:main_results}

\noindent\textbf{Class-conditional generation.}
Table~\ref{tab:imagenet_comparison} shows that NTI achieves strong performance with a standard raster-order AR backbone. On the same AR-L backbone, NTI-L reduces FID from $3.02$ to $1.67$ and improves IS from $256.20$ to $312.72$. Figure~\ref{fig:scaling_effect} further shows consistent gains across NTI-S/B/L as model size increases. NTI-L also outperforms reAR-L, RAR-L, LlamaGen-L, and BitDance-B-1x in FID while retaining a standard discrete tokenizer and raster-order generation. With only $461$M parameters and $400$ training epochs, it remains competitive with substantially larger or more expensive state-of-the-art generative models, showing that NTI can markedly improve visual AR generation without specialized architectures, tokenizers, or decoding strategies.

\noindent\textbf{Generalization.}
We further evaluate NTI on GigaTok~\citep{xiong2025gigatok} and AliTok~\citep{wu2025alitok}, keeping their original tokenizers and generators unchanged and only adding NTI training. NTI improves FID from $3.94$ to $3.66$ on GigaTok and from $1.61$ to $1.51$ on AliTok, demonstrating that NTI generalizes across different visual AR configurations.

\noindent\textbf{Computational cost.}
NTI introduces little additional computation. On 8 NVIDIA A100 GPUs, the estimated per-epoch training times of vanilla AR, NTI, and online two-pass construction are $11.62$, $12.80$, and $22.90$ minutes, respectively. The NTI estimate includes the cost of periodic context construction using a frozen snapshot, averaged over the epochs between refreshes. Thus, NTI adds only about $10\%$ training overhead, whereas online two-pass training nearly doubles the cost. NTI does not modify the inference pipeline; NTI-S/B/L achieve throughputs of $20.709/17.519/13.957$ images/s.

\subsection{Ablation Studies}
\label{sec:ablation}

We ablate the key design choices of NTI, including CRI, SCA, and the training hyperparameters.

\begin{wraptable}{r}{0.42\columnwidth}
    \centering
    \vspace{-0.7em}
    \caption{\textbf{Ablation studies of NTI.}}
    \label{tab:nti_ablation}
    \vspace{-0.45em}

    \footnotesize
    \setlength{\tabcolsep}{3.5pt}
    \renewcommand{\arraystretch}{0.98}

    \begin{tabularx}{\linewidth}{@{}Xr@{}}
        \toprule
        Ablation setting & FID $\downarrow$ \\
        \midrule

        \multicolumn{2}{@{}l@{}}{\textit{Training strategy (400 epochs)}} \\
        \quad SCA + GT supervision & 2.11 \\
        \quad GT context + context-independent & 2.10 \\
        \quad \textbf{NTI (SCA + CRI)} & \textbf{2.04} \\
        \quad Scheduled sampling + CRI & 2.03 \\

        \midrule
        \multicolumn{2}{@{}l@{}}{\textit{Loss weight (80 epochs)}} \\
        \quad $\lambda := 0.5$ & 3.63 \\
        \quad $\lambda := 1.0$ & 3.61 \\

        \midrule
        \multicolumn{2}{@{}l@{}}{\textit{Peak replacement probability (400 epochs)}} \\
        \quad $\rho_{\max}=0.10$ & 2.09 \\
        \quad $\rho_{\max}=0.20$ & 2.05 \\
        \quad $\rho_{\max}=0.30$ & \textbf{2.04} \\
        \quad $\rho_{\max}=0.40$ & 2.09 \\

        \midrule
        \multicolumn{2}{@{}l@{}}{\textit{Codebook temperature (80 epochs)}} \\
        \quad $\tau := 0.05$ & \textbf{3.55} \\
        \quad $\tau := 0.10$ & 3.63 \\

        \bottomrule
    \end{tabularx}

    \vspace{-0.6em}
\end{wraptable}

\noindent\textbf{CRI and SCA.}
We ablate CRI and SCA using NTI-S, with all variants trained for $400$ epochs. For context-independent supervision, we follow prior work that constructs soft targets from codebook similarity~\citep{hu2025improving,tang2025exploiting}. As shown in Table~\ref{tab:nti_ablation}, this strategy achieves an FID of $2.10$ under the GT context, while SCA with GT supervision yields $2.11$. These results show that neither context-independent imagination nor model-generated contexts alone are sufficient. Under the same SCA contexts, CRI reduces FID from $2.11$ to $2.04$, demonstrating the importance of regulating imagination according to the current context. Online scheduled sampling with CRI achieves $2.03$ FID, only $0.01$ lower than NTI, showing that SCA preserves nearly all the benefit of online context construction while eliminating its additional prediction pass.

\noindent\textbf{Loss weight.}
As shown in Table~\ref{tab:nti_ablation}, the loss weight has little impact on generation quality, with $\lambda=0.5$ and $1.0$ yielding comparable performance. This limited sensitivity may be partly attributed to the scale-free property of AdamW, whose adaptive updates are relatively insensitive to gradient rescaling~\citep{loshchilov2019decoupled,zhuang2022understanding}. We therefore use $\lambda=0.5$ for all experiments.

\noindent\textbf{Replacement probability.}
We ablate the peak replacement probability $\rho_{\max}$ in
$\rho\sim\mathcal{U}(0,\rho_{\max}(s))$, which controls the upper bound of context replacement in SCA. Using NTI-S trained for $400$ epochs, increasing the peak $\rho_{\max}$ from $0.10$ to $0.30$ improves FID from $2.09$ to $2.04$, while further increasing it to $0.40$ degrades FID to $2.09$ (Table~\ref{tab:nti_ablation}). This shows that sufficiently exposing the model to off-GT contexts is beneficial, whereas excessive replacement harms generation quality. We therefore use a peak $\rho_{\max}=0.30$.

\noindent\textbf{Codebook temperature.}
We further study the codebook temperature $\tau$ in Equation~\ref{eq:token_affinity}, which controls the concentration of the affinity over non-GT codebook candidates. A smaller $\tau$ emphasizes tokens that are more similar to the GT token, whereas a larger $\tau$ produces a broader candidate distribution. Using NTI-S trained for $80$ epochs, $\tau=0.05$ achieves an FID of $3.55$, compared with $3.63$ for $\tau=0.10$ (Table~\ref{tab:nti_ablation}). We therefore set $\tau=0.05$ in all experiments.

\section{Related Work}
\label{sec:related_work}

\noindent\textbf{Visual autoregressive generation.}
Early visual AR models perform sequential prediction directly over image elements, conditioned on preceding context~\citep{parmar2018image}. To reduce the cost of pixel-level modeling, tokenizers compress images into discrete code sequences~\citep{oord2017neural,esser2021taming}, enabling scalable causal Transformers for image generation~\citep{sun2024autoregressive}. Recent work has improved components of this pipeline. Tokenizer design has progressed toward more compact and scalable representations~\citep{yu2024image,xiong2025gigatok}, flexible one-dimensional sequences and causality-aligned dependencies~\citep{bachmann2025flextok,wu2025alitok}, and larger discrete vocabularies with sub-token prediction~\citep{luo2024openmagvit2}. Other studies relax the fixed raster order through randomized sequence modeling~\citep{yu2025randomized,pang2025randar}, while alternative formulations replace discrete next token prediction with continuous representations~\citep{li2024autoregressive}, next-scale prediction~\citep{tian2024visual}, or more flexible prediction units~\citep{ren2025beyond}. In contrast, NTI keeps the tokenizer, generation order, and prediction unit unchanged, and instead revisits the supervision of standard next token prediction.

\noindent\textbf{Exposure bias.}
Exposure bias has long been studied in autoregressive sequence modeling, where scheduled sampling exposes the model to its own predictions during training while retaining the GT token as supervision~\citep{bengio2015scheduled}. Similar strategies have been adopted in visual AR models: RQ-Transformer introduces stochastic sampling during training~\citep{lee2022autoregressive}, IQ-VAE mixes GT and predicted tokens through Gumbel-softmax~\citep{zhan2022autoregressive}, and reAR introduces noisy contexts while continuing to predict the GT next token~\citep{he2026rear}. More recently, autoregressive video generation methods have reduced the training--inference gap through inference-like contexts or self-generated rollouts~\citep{zhou2025taming,huang2025selfforcing}. These approaches reduce the training--inference mismatch through different context-construction and supervision mechanisms. NTI instead focuses on the supervision under off-GT contexts: an off-GT context can still represent a plausible visual state, and its next token need not be restricted to the GT token. NTI therefore regulates plausible non-GT supervision according to the current context, enabling the model to explore valid continuations beyond the observed GT sequence.

\noindent\textbf{Supervision beyond the GT token.}
Beyond-GT supervision has been explored in discrete autoregressive modeling. OCD derives context-dependent valid next token targets from optimal completions of generated prefixes using dynamic programming~\citep{sabour2019optimal}, while DIST$^2$Loss replaces one-hot targets with distance-aware distributions defined by token metrics~\citep{chung2026teaching}. In visual AR generation, IAR constructs cluster-oriented supervision from codebook similarity~\citep{hu2025improving}, and DCPE extracts discriminative priors from the visual codebook~\citep{tang2025exploiting}. However, DIST$^2$Loss, IAR, and DCPE determine non-GT supervision from token relationships without adapting it to the current AR context. NTI instead treats plausible non-GT next tokens as context-dependent valid continuations and regulates their supervision according to the current context. Additional discussion is provided in Appendix~\ref{app:related_discussion}.

\section{Conclusion}
\label{sec:conclusion}

In this paper, we identify a key limitation of visual AR training: standard supervision treats the GT token as the sole target, even though plausible non-GT next tokens can lead to valid continuations. To address this, we propose \textbf{Next Token Imagination (NTI)}, which reframes next token prediction from GT-only supervision to other plausible continuations. With CRI and SCA, NTI substantially improves standard visual AR models, and generalizes across different tokenizer and generator designs. We hope this work encourages future visual AR models to move beyond the visual space covered by the training data and explore a broader space of plausible visual content.

\subsection*{AI use statement}
We used generative AI tools, including ChatGPT and Codex,
to assist with research ideation, experimental design,
code development and debugging, literature retrieval,
and the preparation of figures and tables. We also used
these tools to draft and revise portions of the manuscript
and improve writing clarity. The authors reviewed and
revised the AI-assisted content, checked references against
the original sources, and tested AI-assisted code.
We take full responsibility for the final content of this
work, including its text, methods, experimental results,
and conclusions.

\bibliography{iclr2027_conference}
\bibliographystyle{iclr2027_conference}

\appendix

\section{Additional Experimental Details}
\label{app:experimental_details}

\paragraph{Code and checkpoints.}
Our training and evaluation code, together with the NTI-S/B/L inference
checkpoints, is publicly available at
\url{https://github.com/JackRuihang/NTI}.
The repository includes model configurations and instructions for sampling
and computing FID and IS.

\subsection{Dataset, Backbone, and Optimization}
\label{app:training_details}

\paragraph{Dataset and evaluation.}
We evaluate class-conditional image generation on ImageNet-1K
at $256\times256$ resolution~\citep{deng2009imagenet}.
For each evaluation, we generate $50{,}000$ images and compute
FID and IS following the ADM evaluation
protocol~\citep{dhariwal2021diffusion,heusel2017gans,salimans2016improved}.
We use the same ImageNet-$256$ reference archive,
\texttt{VIRTUAL\_imagenet256\_labeled.npz},
for all evaluations.

\paragraph{Tokenizer and generator.}
For the standard raster-order experiments, we use a frozen
MaskGIT-VQGAN tokenizer~\citep{chang2022maskgit} with a
downsampling factor of $16$, producing a $16\times16$ token grid
and a sequence length of $256$.
The tokenizer has a codebook of $1{,}024$ entries with an
embedding dimension of $256$.
NTI-S/B/L use $20/24/24$ causal Transformer layers with hidden
dimensions of $768/768/1024$, respectively.
All variants use $16$ attention heads, learned positional
embeddings, class-conditioned adaptive layer normalization,
query--key normalization, an MLP expansion ratio of $4$, and
dropout of $0.1$ in the attention and feed-forward layers.

\paragraph{Optimization.}
We train the standard raster-order models for $250{,}000$
optimization steps (approximately $400$ epochs) with a global
batch size of $2{,}048$.
We use AdamW~\citep{loshchilov2019decoupled} with
$(\beta_1,\beta_2)=(0.9,0.96)$ and weight decay $0.03$.
The learning rate is linearly warmed up to $4\times10^{-4}$
during the first $62{,}500$ steps and then decayed to
$10^{-5}$ with a cosine schedule.
Training uses bfloat16 mixed precision and gradient clipping
with a maximum norm of $1.0$.
Class labels are dropped with probability $0.1$ for
classifier-free guidance~\citep{ho2022classifierfree}.

\subsection{Snapshot-Based Context Construction}
\label{app:sca_implementation}

\paragraph{Replacement schedule.}
Let $s\in[0,1]$ denote normalized training progress and
$\rho_{\max}$ denote the peak replacement probability.
We use a cosine rise-and-decay schedule:
\begin{equation}
\rho_{\max}(s)=
\begin{cases}
\displaystyle
\frac{\rho_{\max}}{2}
\left[
1-\cos\left(\pi\frac{s}{0.25}\right)
\right],
& 0\leq s<0.25, \\[6pt]
\displaystyle
\frac{\rho_{\max}}{2}
\left[
1+\cos\left(\pi\frac{s-0.25}{0.55}\right)
\right],
& 0.25\leq s<0.8, \\[6pt]
0,
& 0.8\leq s\leq1.
\end{cases}
\label{eq:app_replacement_schedule}
\end{equation}
We set the peak replacement probability to
$\rho_{\max}=0.3$ by default.
The schedule reaches its peak at $25\%$ of training and
decreases to zero at $80\%$, after which only GT sequences
are used.
For each sequence, we sample
$\rho\sim\mathcal{U}(0,\rho_{\max}(s))$.
The CRI affinity temperature is set to $\tau=0.05$ unless
otherwise specified.

\paragraph{Periodic context construction.}
Context construction starts at step $10{,}000$ and is
refreshed every $10{,}000$ steps until step $200{,}000$,
corresponding to approximately $16$ epochs between refreshes.
At each refresh, a frozen snapshot computes predictions
under GT contexts, constructs SCA contexts.

\paragraph{Single-pass optimization.}
GT sequences and cached SCA records are mixed within each
training batch and processed in a single forward pass of the
trainable generator.
Snapshot computations are performed only during periodic
context construction.
The cached SCA contexts and CRI targets remain fixed between
successive refreshes.

\paragraph{NTI loss normalization.}
We use $\mathbb{E}_{\mathrm{SCA}}$ in Equation~\ref{eq:nti_loss} to denote the expectation over
\[
(\mathbf{z},y)\sim\mathcal{D}_{\mathrm{train}},\qquad
\rho\sim\mathcal{U}(0,\rho_{\max}(s)),\qquad
\widetilde{\mathbf{z}}\sim
\mathcal{P}_{\mathrm{SCA}}
(\cdot\mid\mathbf{z},y,\rho;\bar{\theta}).
\]
The NTI loss is normalized by the total number of valid NTI positions rather than independently within each sequence. With $m_t=\mathbf{1}[\widetilde{\mathbf{z}}_{<t}\neq\mathbf{z}_{<t}]$, all positions with $m_t=1$ contribute equally to $\mathcal{L}_{\mathrm{NTI}}$. When no valid NTI positions are present, only the standard AR objective is applied.

\paragraph{Effective supervision weight.}
We use a base weight of $\lambda=0.5$ unless otherwise
specified.
The effective NTI supervision weight follows the replacement
schedule:
\begin{equation}
\lambda(s)
=
\lambda\frac{\rho_{\max}(s)}{\rho_{\max}}.
\label{eq:app_effective_weight}
\end{equation}
In the mixed-record implementation, this relative weighting
is realized through record sampling.
With the GT record weight set to $1$, the probability of
sampling an SCA record is
\begin{equation}
\pi(s)
=
\frac{\lambda(s)}{1+\lambda(s)}.
\label{eq:app_record_mixing}
\end{equation}
We evaluate $\pi(s)$ at the midpoint of each refresh interval.

\subsection{Sampling and Efficiency Evaluation}
\label{app:sampling_details}

\paragraph{Sampling settings.}
The standard raster-order generator samples $256$ tokens
and decodes them with the frozen tokenizer.
We use a sampling temperature of $1.0$ with no top-$k$
or nucleus truncation.
Following prior work~\citep{gao2023masked,he2026rear},
we use a power-cosine classifier-free guidance schedule.
At zero-based token position $j$, the guidance strength is
\begin{equation}
w_j
=
1+\frac{w-1}{2}
\left[
1-\cos\left(\pi(j/N)^\gamma\right)
\right],
\qquad j=0,\ldots,N-1,
\label{eq:app_cfg_schedule}
\end{equation}
where $N=256$, $w$ is the guidance-scale parameter, and
$\gamma$ controls its progression.
NTI-S and NTI-B use
$(w,\gamma)=(22,2.75)$ and $(14.5,2.25)$, respectively.
For the NTI-L result reported in
Table~\ref{tab:imagenet_comparison}, we use
$(w,\gamma)=(14,1.75)$.

\paragraph{Training-budget curves.}
Figure~\ref{fig:scaling_effect} evaluates checkpoints at
$50$K, $100$K, $150$K, $200$K, and $250$K optimization steps,
corresponding to approximately $80$, $160$, $240$, $320$, and
$400$ epochs.
The first four checkpoints are evaluated with sampling seed
$42$.
For the final checkpoint, we report the average over sampling
seeds $42$, $142$, and $242$.

\paragraph{Sampling throughput.}
We measure sampling throughput on a single NVIDIA
A100-SXM4-80GB GPU with batch size $128$.
The timed region includes token generation and tokenizer
decoding, but excludes image-file writing.
Sampling uses KV caching and FP32 inference without autocast
or model compilation.
We discard three warm-up batches and run three timing
repetitions, each with at least ten batches and $30$ seconds
of sampling.
CUDA synchronization is applied at the timing boundaries.

\paragraph{Training-cost accounting.}
We estimate per-epoch training time by accounting for the
optimization cost and periodic SCA construction cost.
For NTI, the latter includes context construction
with a frozen snapshot and is averaged over the epochs between
successive refreshes.
The online comparator performs two-pass context construction
at each optimization step.

\section{Analysis Details of Next Token Imagination}
\label{app:nti_analysis_details}

This section provides additional details for the two analysis
experiments in Section~\ref{sec:understanding}. We describe the
controlled comparisons, evaluation metrics, and interpretation of the results, following the GTR and GTS definitions in the main text.

\subsection{Feasibility of Next Token Imagination}
\label{app:nti_feasibility_details}

\noindent\textbf{Controlled comparisons.}
The experiment in Figure~\ref{fig:experiment1} examines how deviations in the context affect subsequent generation. We construct perturbed contexts from GT contexts within the context window defined in Section~\ref{sec:understanding}. We consider two settings: varying GTR at fixed GTS, and varying GTS while fixing GTR and the replacement positions. Non-GT replacements are selected according to their codebook cosine similarity to the corresponding GT tokens, separating the extent of context replacement from the similarity of the replaced tokens.

Starting from each perturbed context, we autoregressively generate the remaining tokens. We additionally generate a continuation from the corresponding GT context using the same source, target position, and sampling randomness. Let $I_{\mathrm{pert}}$ and $I_{\mathrm{GT}}$ denote the decoded images generated from the perturbed and GT contexts, respectively. The following image-level metrics compare these paired generations.

\noindent\textbf{Metric conventions.}
For all analyzed positions, we use a non-empty context window with $L>0$. When $\mathcal{A}=\emptyset$, the perturbed and GT contexts coincide within $\mathcal{W}_t$; in this case, we set $\operatorname{GTR}=1$ and $\operatorname{GTS}=1$. Otherwise, GTR and GTS follow the definitions in Section~\ref{sec:understanding}.

\noindent\textbf{Perceptual deviation.}
We use LPIPS~\citep{zhang2018unreasonable} to measure the perceptual difference between the paired generations:
\begin{equation}
\label{eq:app_lpips_pair}
\begin{aligned}
\operatorname{LPIPS}(I_{\mathrm{pert}},I_{\mathrm{GT}})
=
\sum_{\ell}\frac{1}{H_{\ell}W_{\ell}}
\sum_{h,w}
\left\|
\mathbf{w}_{\ell}\odot
\left(
\widehat{\mathbf{f}}_{\ell}(I_{\mathrm{pert}})_{hw}
-
\widehat{\mathbf{f}}_{\ell}(I_{\mathrm{GT}})_{hw}
\right)
\right\|_2^2,
\end{aligned}
\end{equation}
where $\widehat{\mathbf{f}}_{\ell}$ denotes the channel-normalized feature map at layer $\ell$ of the perceptual network, $H_{\ell}\times W_{\ell}$ is its spatial resolution, and $\mathbf{w}_{\ell}$ contains the learned calibration weights. Lower LPIPS indicates greater perceptual similarity between the generations from the perturbed and GT contexts.

\noindent\textbf{Pixel-wise and structural similarity.}
We supplement LPIPS with PSNR and SSIM~\citep{wang2004image}. For RGB images with pixel values in $[0,1]$, PSNR is defined as
\begin{equation}
\operatorname{MSE}
=
\frac{
\left\|I_{\mathrm{pert}}-I_{\mathrm{GT}}\right\|_2^2
}{3HW},
\qquad
\operatorname{PSNR}
=
10\log_{10}\frac{1}{\operatorname{MSE}}.
\end{equation}
SSIM measures local structural similarity. For corresponding image patches $u$ and $v$, it is given by
\begin{equation}
\operatorname{SSIM}(u,v)
=
\frac{
(2\mu_u\mu_v+C_1)(2\sigma_{uv}+C_2)
}{
(\mu_u^2+\mu_v^2+C_1)
(\sigma_u^2+\sigma_v^2+C_2)
},
\end{equation}
where $\mu$, $\sigma^2$, and $\sigma_{uv}$ denote local means, variances, and covariance, respectively. The positive constants $C_1$ and $C_2$ stabilize the ratios. We average local scores over spatial positions and color channels. Higher PSNR and SSIM indicate greater pixel-wise and structural similarity between the generations from the perturbed and GT contexts.

\noindent\textbf{Paired DINO distance.}
To measure feature-level deviation, we use the normalized global DINOv2 feature~\citep{oquab2024dinov2}, $\phi(I)=f_{\mathrm{DINO}}(I)/\|f_{\mathrm{DINO}}(I)\|_2$, and compute
\begin{equation}
\label{eq:app_paired_dino}
d_{\mathrm{pair}}(I_{\mathrm{pert}},I_{\mathrm{GT}})
=
1-\phi(I_{\mathrm{pert}})^\top\phi(I_{\mathrm{GT}}).
\end{equation}
A smaller distance indicates greater feature similarity between the paired generations. Paired DINO distance measures deviation from the corresponding GT-context generation, whereas the Class-DINO score used in Experiment 2 measures similarity to real images from the target class.

\noindent\textbf{Compatibility with the GT suffix.}
In addition to the image-level metrics, we evaluate the likelihood of a fixed GT suffix under the GT and perturbed contexts. For a context $h$ at position $t$, we define
\begin{equation}
\mathcal{E}(h)
=
-\frac{1}{H_{\mathrm{suf}}}
\sum_{j=t}^{t+H_{\mathrm{suf}}-1}
\log p_{\theta}(z_j\mid h,z_{t:j-1},y),
\qquad H_{\mathrm{suf}}=32,
\end{equation}
where $z_{t:j-1}$ is empty when $j=t$, and each suffix token is evaluated conditioned on the preceding GT suffix tokens. Let $\mathbf{z}^{\mathrm{pert}}_{<t}$ denote the perturbed context. We report
\begin{equation}
\Delta\operatorname{NLL}
=
\mathcal{E}(\mathbf{z}^{\mathrm{pert}}_{<t})
-
\mathcal{E}(\mathbf{z}_{<t}).
\end{equation}
A larger $\Delta\operatorname{NLL}$ indicates lower compatibility between the perturbed context and the GT suffix.

\begin{figure}[t]
    \centering
    \includegraphics[width=0.82\linewidth]{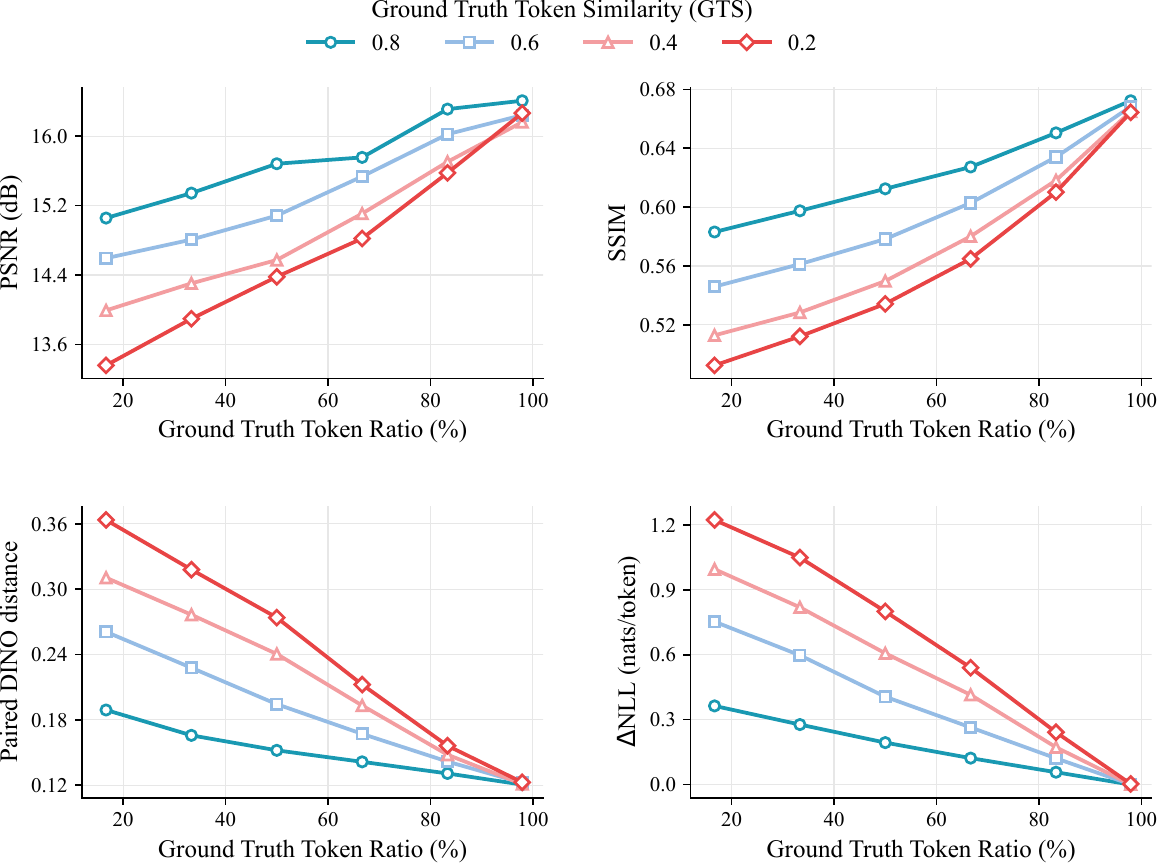}
    \caption{\textbf{Additional measurements for Experiment 1.}
    Colors indicate requested GTS levels. Each curve connects the
    same six measured GTR levels used in the main-text visualization,
    without smoothing. PSNR, SSIM, and paired DINO distance compare
    perturbed-context generations with their paired GT-context
    generations. $\Delta$NLL measures the change in negative
    log-likelihood of a fixed GT suffix.}
    \label{fig:app_nti_metrics}
\end{figure}

\noindent\textbf{Results and interpretation.}
Figure~\ref{fig:app_nti_metrics} shows that lower GTR generally reduces PSNR and SSIM while increasing paired DINO distance and $\Delta$NLL. Under stronger context perturbations, higher GTS generally preserves greater image similarity and yields a smaller increase in GT-suffix NLL.

Together with the decoded examples in Figure~\ref{fig:experiment1}, these results show that larger context deviations increasingly affect subsequent generation, while perturbed contexts can still lead to coherent generations beyond the GT sequence. This further supports the feasibility of NTI beyond the GT sequence.

\subsection{Regulation of Next Token Imagination}
\label{app:nti_regulation_details}

\noindent\textbf{Class-conditional semantic evaluation.}
The experiment in Figure~\ref{fig:experiment2} evaluates the class-conditional semantic similarity of the resulting images using DINOv2~\citep{oquab2024dinov2}. Let $\phi(I)$ denote the normalized DINOv2 feature defined above, and let $\mathcal{R}_y=\{R_{y,1},\ldots,R_{y,M}\}$ be a reference set of real images from target class $y$. The reference set excludes the source images used in the analysis. Let $\mathcal{N}_k(I,y)$ denote the indices of the $k$ reference images with the highest cosine similarity to $I$. We define the Class-DINO score as
\begin{equation}
\label{eq:app_class_dino}
S_y(I)
=
\frac{1}{k}
\sum_{j\in\mathcal{N}_k(I,y)}
\phi(I)^\top\phi(R_{y,j}),
\qquad M=20,\quad k=5.
\end{equation}
A higher score indicates greater feature similarity to real images from the target class.

\noindent\textbf{GT versus non-GT continuations.}
In Figure~\ref{fig:experiment2}(a), we hold the generated context and target class fixed and compare continuations obtained by selecting the GT token or a non-GT token at the next position. The remaining tokens are generated using the same continuation protocol, and the decoded images are evaluated using Equation~\ref{eq:app_class_dino}. Comparable or higher Class-DINO scores for non-GT continuations show that the GT token is not the only choice that can preserve the target-class semantics under a generated context.

\noindent\textbf{Fixed versus context-adaptive selection.}
Figure~\ref{fig:experiment2}(b) compares fixed and context-adaptive selection of non-GT tokens. Fixed selection uses the same non-GT token across different contexts, whereas context-adaptive selection reselects the token according to the current context by combining codebook cosine similarity with the model prediction. The resulting continuations are evaluated using the same Class-DINO score. This comparison measures how the suitability of a non-GT token changes with the current context.

\noindent\textbf{Relation to the plausible visual space.}
Class-DINO measures semantic similarity to real images from the target class. Together with the decoded examples in Figure~\ref{fig:experiment2}, the results show that plausible continuations can extend beyond the observed GT sequence and that suitable non-GT targets depend on the current context.

\section{Qualitative Evidence of Expanded Visual Space}
\label{app:expanded_visual_space}

The controlled analyses in Appendix~\ref{app:nti_analysis_details} characterize deviations from the GT sequence through pixel, perceptual, feature, and likelihood spaces. We further examine the generated images directly to investigate whether the broader supervision introduced by NTI translates into a broader range of plausible visual realizations at inference.

\begin{figure*}[t]
    \centering
    \setlength{\tabcolsep}{3pt}
    \renewcommand{\arraystretch}{0.9}

    \begin{tabular}{cc}
        \textbf{AR-L} & \textbf{NTI-L} \\[2pt]

        \multicolumn{2}{c}{\textit{Volcano}} \\[-1pt]
        \includegraphics[width=0.48\textwidth]{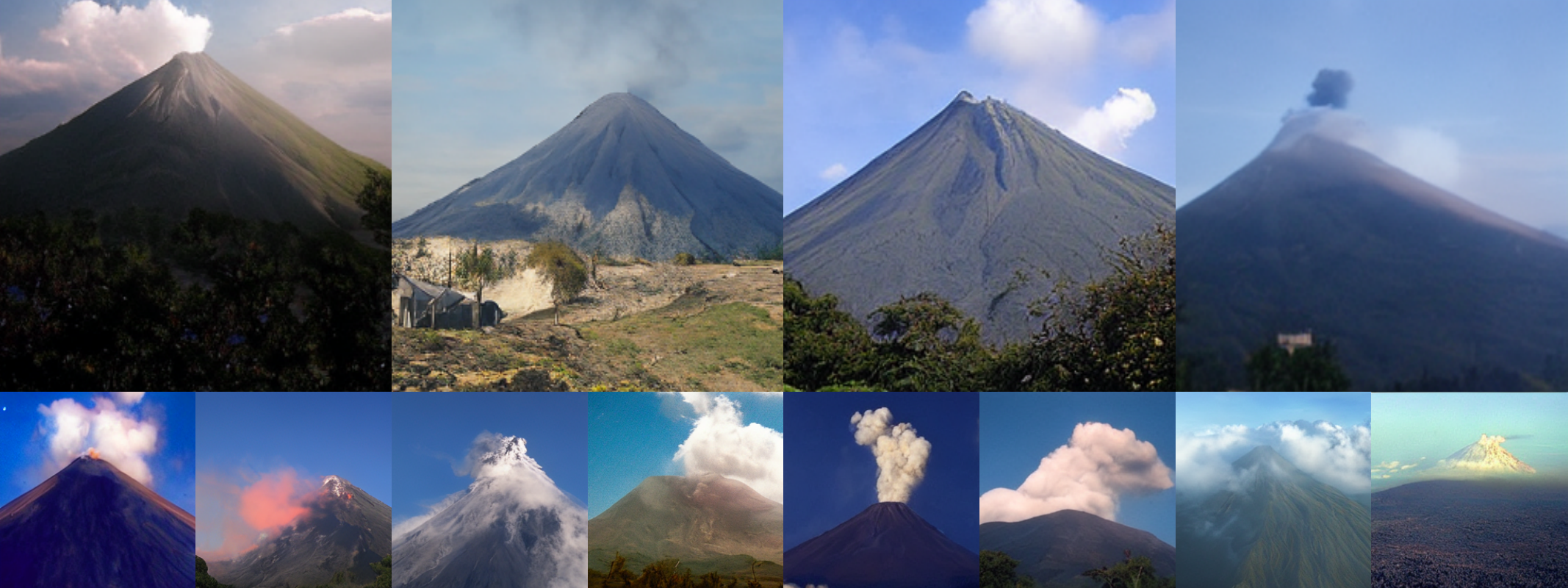} &
        \includegraphics[width=0.48\textwidth]{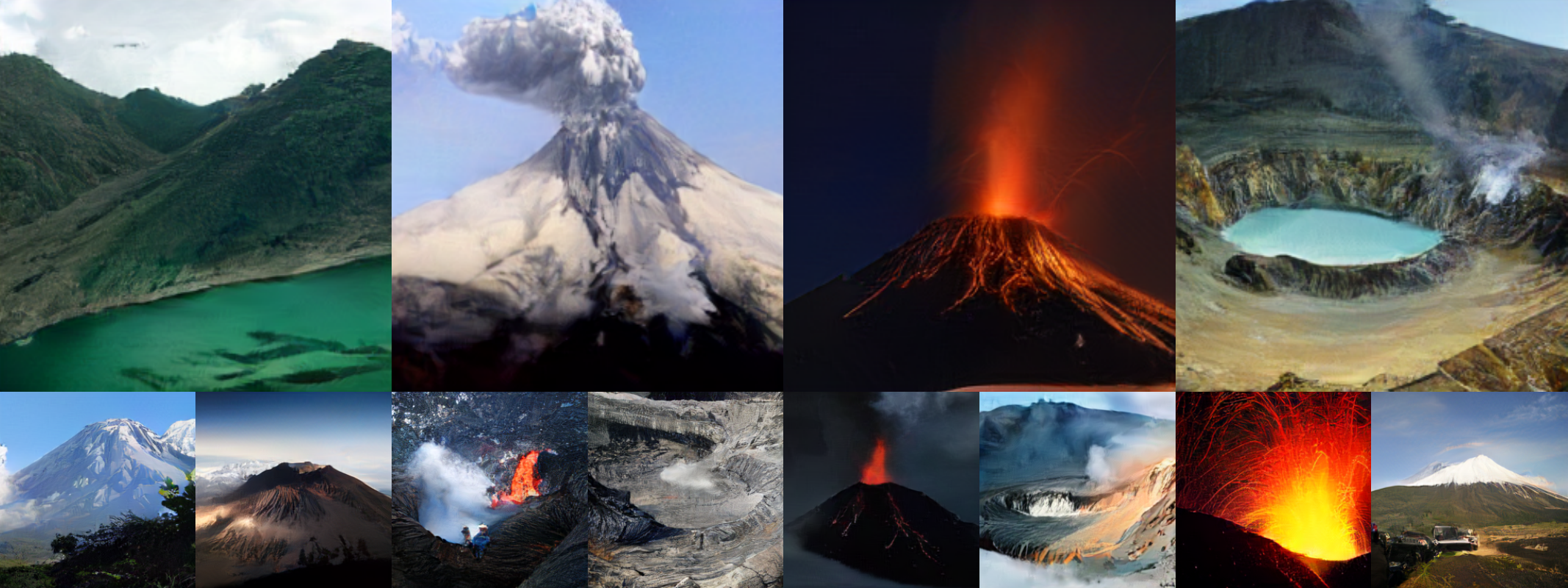} \\[3pt]

        \multicolumn{2}{c}{\textit{Guitar pick}} \\[-1pt]
        \includegraphics[width=0.48\textwidth]{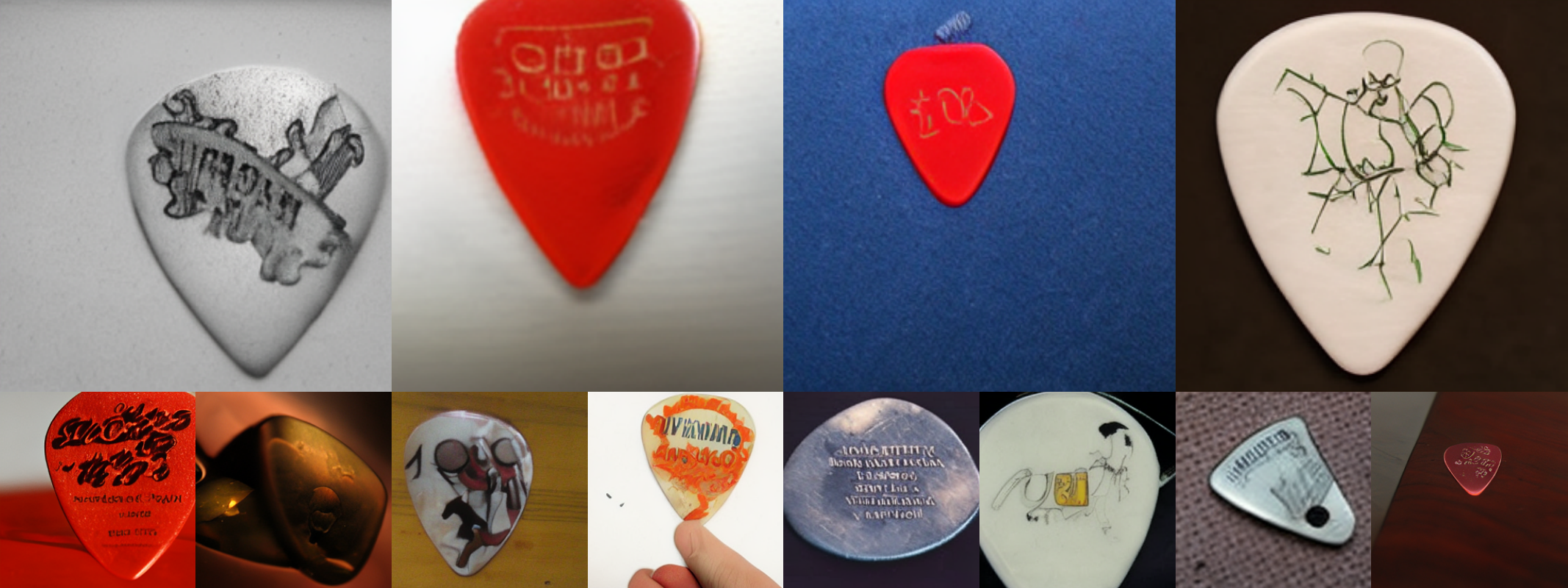} &
        \includegraphics[width=0.48\textwidth]{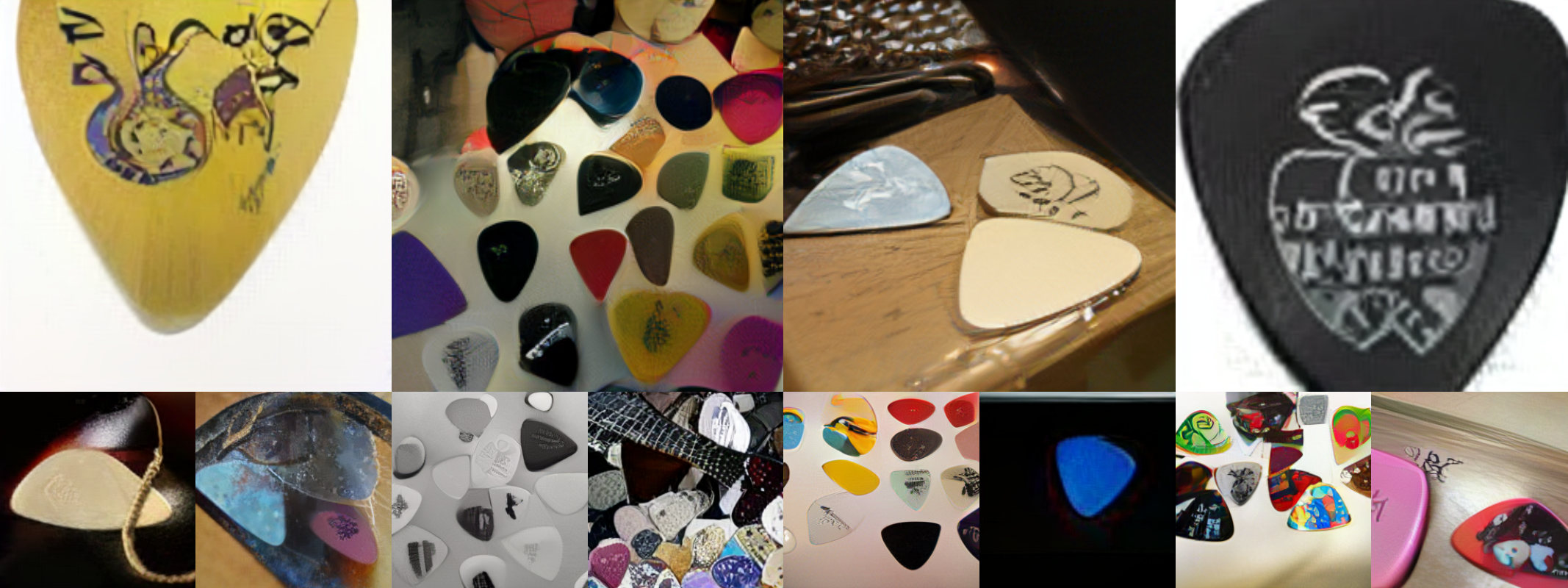} \\[3pt]

        \multicolumn{2}{c}{\textit{Hippopotamus}} \\[-1pt]
        \includegraphics[width=0.48\textwidth]{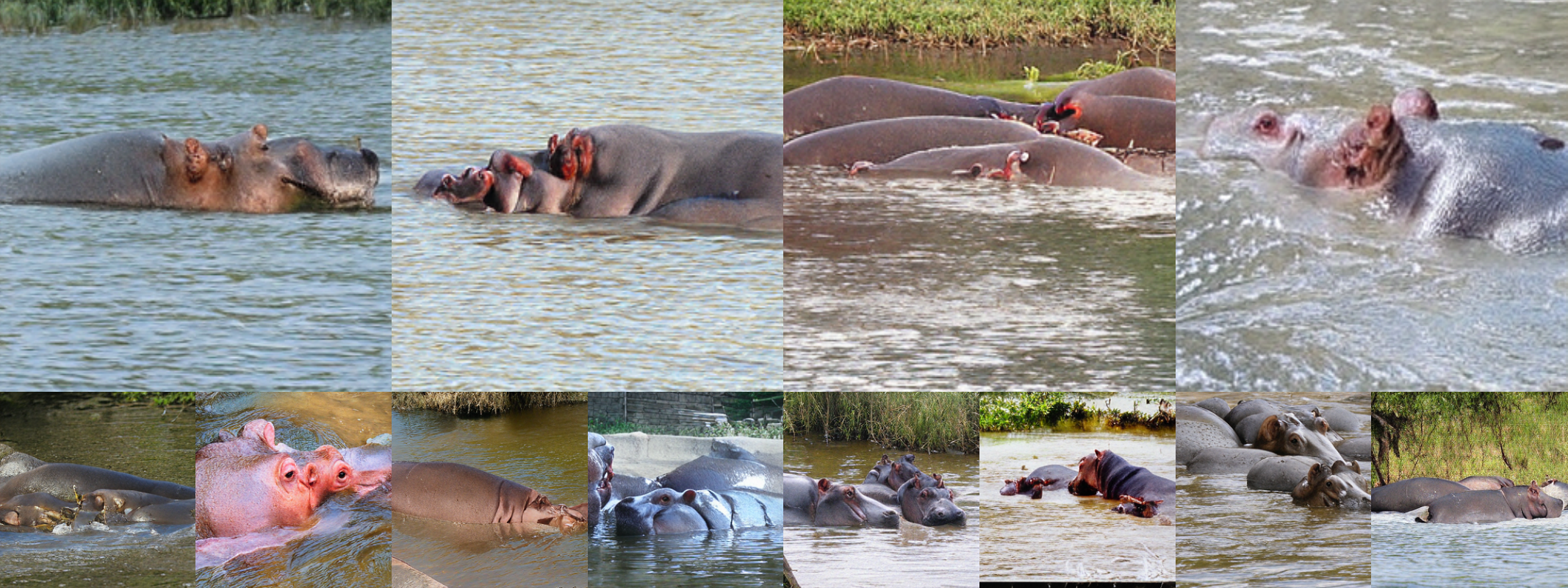} &
        \includegraphics[width=0.48\textwidth]{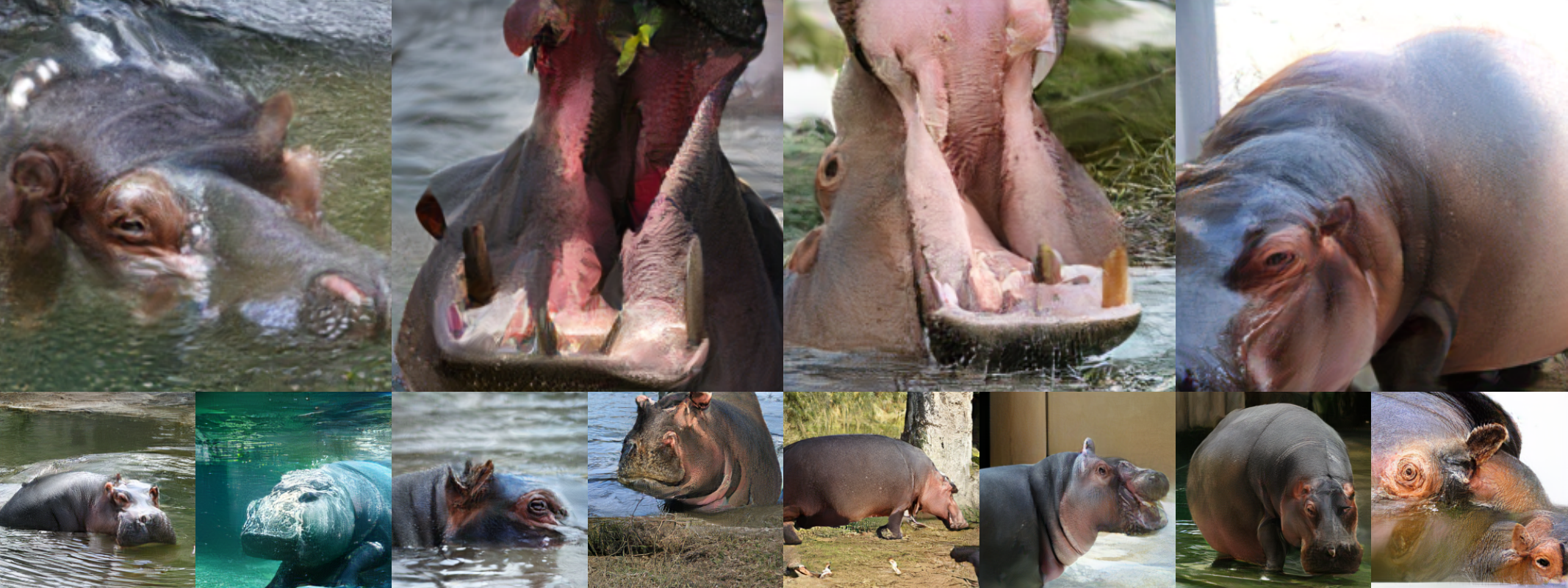} \\[3pt]

        \multicolumn{2}{c}{\textit{Wallaby}} \\[-1pt]
        \includegraphics[width=0.48\textwidth]{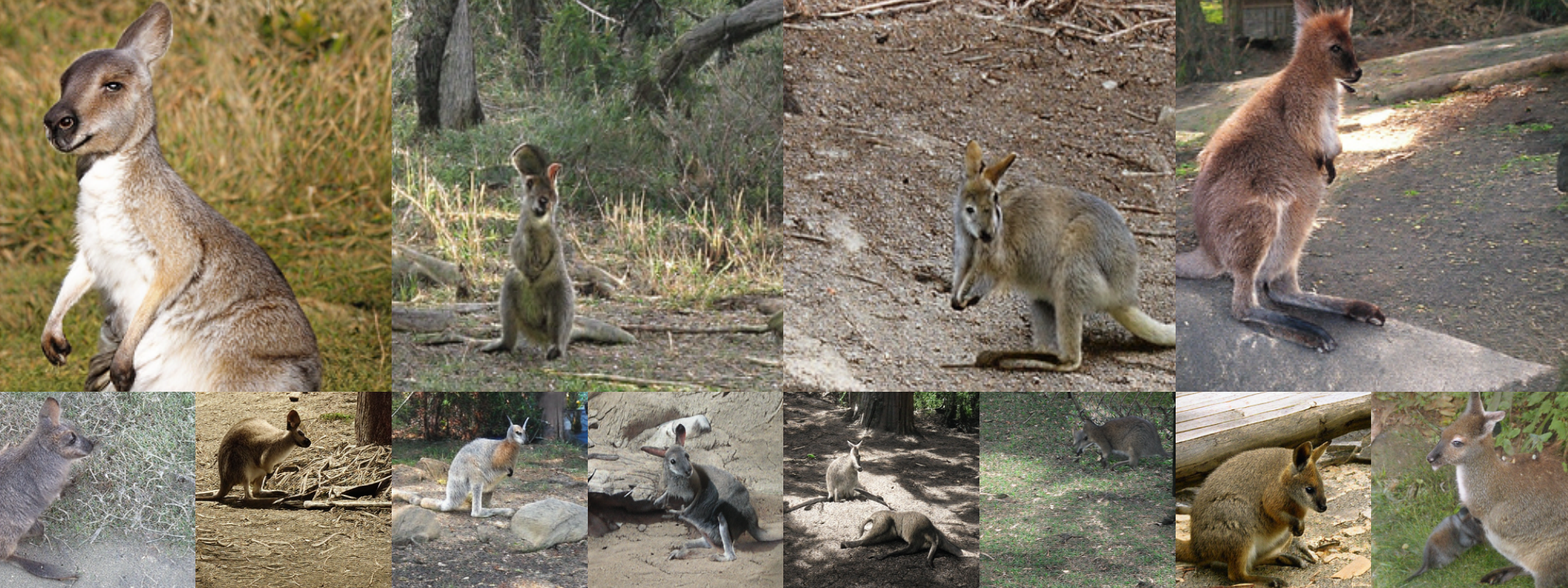} &
        \includegraphics[width=0.48\textwidth]{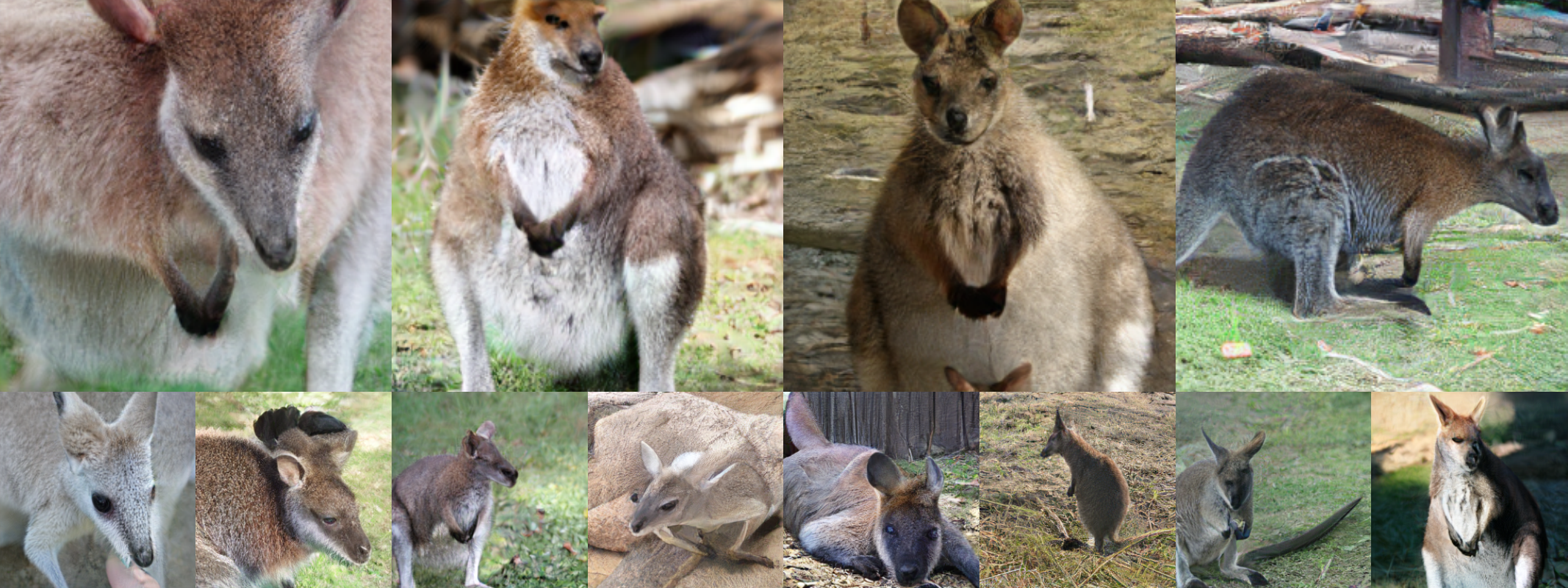}
    \end{tabular}

    \caption{\textbf{NTI expands the range of plausible visual realizations.}
    Compared with AR-L, NTI-L generates substantially greater variation in visual appearance, composition, pose, and viewpoint across the same ImageNet classes. Volcanoes and guitar picks exhibit broader morphological and compositional variation, while hippopotamuses and wallabies show markedly richer pose and viewpoint diversity. These results complement the controlled analyses in Appendix~\ref{app:nti_analysis_details}, showing that the broader supervision introduced by NTI translates into a broader visual space at inference.}
    \label{fig:expanded_visual_space}
\end{figure*}

Figure~\ref{fig:expanded_visual_space} compares samples generated by AR-L and NTI-L under the same class conditions. The difference is particularly visible in object morphology and composition. For volcanoes, AR-L predominantly generates distant conical mountains with similar eruption patterns, whereas NTI-L produces substantially more varied realizations, including crater lakes, explosive ash plumes, lava eruptions, snow-covered volcanoes, and close-up volcanic structures. For guitar picks, AR-L mainly generates isolated picks with similar centered compositions, while NTI-L exhibits greater variation in appearance, arrangement, viewpoint, and interaction with surrounding objects.

The same effect appears in object pose. AR-L hippopotamuses are frequently generated in similar partially submerged configurations, whereas NTI-L produces markedly different poses and viewpoints, including frontal close-ups, open-mouth poses, and standing or on-land configurations. Wallabies generated by NTI-L likewise span a broader range of viewpoints and poses, including close-up, frontal, side-view, standing, walking, and crouching configurations.

Together with the controlled experiments in Appendix~\ref{app:nti_analysis_details}, these qualitative results show that NTI expands the range of plausible visual realizations produced by a standard visual AR model. The broader supervision encountered during NTI training therefore translates into a broader visual space at inference.

\section{Additional Discussion on Related Work}
\label{app:related_discussion}

This section expands the comparisons in
Section~\ref{sec:related_work}.
We discuss approaches that improve visual AR generation
through generation order, tokenization, alternative
prediction formulations, and additional training supervision.
Across these directions, we focus on how they differ from NTI,
which extends next token supervision beyond the GT token and
regulates non-GT supervision according to the current context.

\subsection{Randomized Generation Order}

\noindent\textbf{RandAR.}
RandAR~\citep{pang2025randar} introduces position-instruction
tokens to specify the spatial locations of tokens to be
generated.
This allows a decoder-only Transformer to generate image
tokens in randomized orders rather than a fixed raster order.
By providing multiple position instructions, RandAR also
supports parallel decoding.
Its approach changes both the sequence organization and the
generation procedure.

\noindent\textbf{RAR.}
RAR~\citep{yu2025randomized} incorporates target-position
information into autoregressive prediction and trains with
randomized token orders.
The degree of randomization decreases during training,
eventually recovering raster-order modeling.
This exposes the generator to different spatial contexts
during training while retaining raster-order autoregressive
generation at inference.

\noindent\textbf{ARPG.}
ARPG~\citep{li2026autoregressive} separates positional queries
from content representations to support randomized parallel
generation.
Its shared KV cache allows previously computed content
representations to be reused when querying multiple positions.
ARPG therefore improves both generation-order flexibility
and parallel decoding efficiency.

\noindent\textbf{Summary.}
Randomized-order methods vary which spatial tokens precede
each prediction in the autoregressive sequence, exposing the
generator to contexts beyond a fixed raster order.
NTI instead retains raster-order generation and modifies the
supervision under a given context by allowing plausible
non-GT next tokens according to the current context.

\subsection{1D and Causality-Aligned Tokenization}

\noindent\textbf{TiTok.}
TiTok~\citep{yu2024image} represents images with compact
one-dimensional token sequences rather than a fixed
two-dimensional token grid.
This substantially reduces the number of visual tokens
required for reconstruction and generation.

\noindent\textbf{GigaTok.}
GigaTok~\citep{xiong2025gigatok} studies the scaling of visual
tokenizers for autoregressive generation.
It introduces semantic regularization with pretrained visual
features and scales the tokenizer to improve both
reconstruction quality and downstream generation.
Its contribution primarily concerns the representation
provided to the generator.

\noindent\textbf{FlexTok.}
FlexTok~\citep{bachmann2025flextok} resamples images into
ordered one-dimensional token sequences of flexible length.
Its coarse-to-fine representation allows different token
budgets to capture different levels of visual detail,
supporting reconstruction and generation at varying
sequence lengths.

\noindent\textbf{AliTok.}
AliTok~\citep{wu2025alitok} addresses the dependency mismatch
between visual tokenization and autoregressive generation.
It uses a bidirectional encoder constrained by a causal
decoder to learn representations that are more compatible
with forward sequence modeling.
A two-stage tokenizer training procedure further improves
reconstruction while preserving this dependency structure.

\noindent\textbf{CaTok.}
CaTok~\citep{chen2026catok} develops a one-dimensional causal
image tokenizer whose token dependencies are explicitly
aligned with autoregressive generation.
It uses a MeanFlow decoder to construct causal representations
over ordered token intervals.

\noindent\textbf{SpectralAR.}
SpectralAR~\citep{huang2025spectralar} constructs ordered
spectral tokens from low- to high-frequency image components.
This produces a causal coarse-to-fine visual sequence for
autoregressive generation.

\noindent\textbf{Hita.}
Hita~\citep{zheng2025holistic} introduces a holistic-to-local
tokenization scheme, placing learned holistic tokens before
local patch tokens and using causal attention to align the
representation with autoregressive generation.

\noindent\textbf{Summary.}
These approaches improve token compactness, representation
quality, or compatibility with causal generation.
NTI instead modifies the supervision used to train the
generator on a given representation.
Table~\ref{tab:generalization} shows that NTI reduces FID
from $3.94$ to $3.66$ on GigaTok and from $1.61$ to $1.51$
on AliTok without changing their tokenizers or generator
architectures.
These results show that tokenizer improvements and
context-regulated supervision are complementary.

\subsection{Alternative Autoregressive Formulations}

\noindent\textbf{MAR.}
MAR~\citep{li2024autoregressive} extends autoregressive image
generation to continuous visual tokens.
It models per-token conditional distributions with a diffusion
loss and further generalizes the formulation to masked
autoregressive generation.
Its prediction mechanism therefore differs from categorical
next token prediction over a discrete codebook.

\noindent\textbf{VAR.}
VAR~\citep{tian2024visual} replaces next token prediction
with next-scale prediction.
Starting from a coarse representation, it progressively
generates finer token maps conditioned on previously generated
scales, with tokens within each scale predicted in parallel.
This changes both the prediction unit and the autoregressive
factorization of the image.

\noindent\textbf{xAR.}
xAR~\citep{ren2025beyond} generalizes the prediction unit from
individual tokens to flexible entities such as cells,
subsamples, scales, or entire images.
It further replaces discrete token classification with
continuous regression using flow matching and introduces
noisy-context training.

\noindent\textbf{Other prediction formulations.}
Other baselines in Table~\ref{tab:imagenet_comparison} adopt
different prediction spaces or generation mechanisms.
MaskBit~\citep{weber2024maskbit} performs masked generation
over binary token representations.
SphereAR~\citep{ke2026hyperspherical} performs continuous-token
autoregressive generation with hyperspherical latents.
BAR~\citep{tang2026bar} learns an end-to-end transformation
of the token basis for autoregressive generation.
BitDance~\citep{ai2026bitdance} uses a binary diffusion head
for token prediction and supports parallel next-patch
generation.

\noindent\textbf{Summary.}
These methods modify the prediction space, prediction unit,
or autoregressive factorization.
NTI instead retains discrete raster-order next token
prediction and constructs a context-regulated supervision
distribution over the existing vocabulary.

\subsection{Training Regularization and Supervision}

\noindent\textbf{REPA.}
REPA~\citep{yu2025repa} improves diffusion Transformer
training by aligning intermediate representations with
features from a pretrained visual encoder.
This provides an additional representation-learning signal
during denoising, acting on intermediate features rather
than directly modifying the output target distribution.

\noindent\textbf{reAR.}
reAR~\citep{he2026rear} improves consistency between the
autoregressive generator and its tokenizer.
It introduces token-wise regularization based on tokenizer
representations and trains the generator under noisy contexts,
while retaining GT next token supervision.

\noindent\textbf{RQ-Transformer.}
RQ-Transformer~\citep{lee2022autoregressive} introduces soft
labeling and stochastic sampling for visual AR training.
Its soft targets are constructed from distances between
codebook embeddings, while stochastic sampling exposes the
generator to different discrete code sequences.

\noindent\textbf{IQ-VAE.}
IQ-VAE~\citep{zhan2022autoregressive} uses reliability-based
Gumbel sampling to mix predicted tokens with GT tokens during
training, reducing the discrepancy between training and
autoregressive inference.

\noindent\textbf{MAGI.}
MAGI~\citep{zhou2025taming} studies exposure bias in masked
autoregressive video generation and introduces training
strategies that expose the model to inference-like conditions.

\noindent\textbf{Token-aware supervision.}
IAR~\citep{hu2025improving} and
DCPE~\citep{tang2025exploiting} exploit codebook structure
and discriminative priors to construct supervision beyond
one-hot GT targets.
DIST$^2$Loss~\citep{chung2026teaching} similarly introduces
distance-aware supervision based on token metrics.
These methods use relationships among tokens to shape the
supervision distribution.

\noindent\textbf{Generated-context supervision.}
OCD~\citep{sabour2019optimal} derives valid next token targets
from optimal continuations of generated prefixes under an
edit-distance objective.
Self Forcing~\citep{huang2025selfforcing} trains autoregressive
video models using self-generated rollouts and sequence-level
distribution matching.
Both consider supervision beyond standard GT-context training,
but operate under different sequence objectives and prediction
spaces.

\noindent\textbf{Summary.}
NTI directly modifies the next token supervision distribution
for visual AR generation.
Codebook affinity identifies plausible non-GT candidates,
while predictions under GT and model-generated contexts
regulate their relative weights and overall supervision
strength.
The resulting supervision therefore adapts to the current
context while retaining the standard discrete visual AR
formulation.

\end{document}